\documentclass{article}
\usepackage[final]{corl_2026} 
\usepackage{ifthen}

\newcommand\arxivmode{false}

\usepackage[numbers]{natbib} 
\usepackage{graphics} 
\usepackage{epsfig} 
\usepackage{times}
\usepackage{amsmath} 
\usepackage{amssymb}  
\usepackage{balance} 
\usepackage{bm} 
\usepackage{tikz}
\usepackage{pgfplots}
\usepackage{wrapfig}
\usepackage[capitalise]{cleveref}
\usepackage{siunitx}
\usepackage{subfig}
\usepackage{booktabs}
\usepackage{multirow}
\usepackage{cuted}
\usepackage{makecell}
\usepackage{soul}
\usepackage{xfrac}
\usepackage{hyperref}

\newcommand{\Sp}{\mathbb{S}}

\DeclareMathOperator*{\argmin}{arg\,min}

\usepackage{xcolor} 

\definecolor{worst}{HTML}{B60909}
\definecolor{mediocre}{HTML}{2170A5}
\definecolor{good}{HTML}{E58328}
\definecolor{best}{HTML}{277318}
\definecolor{yawA}{HTML}{832C94}
\definecolor{yawB}{HTML}{26AC9D}
\definecolor{yawC}{HTML}{ECBB3F}

\newcommand{\best}[1]{{\color{best}#1}}
\newcommand{\good}[1]{{\color{good}#1}}

\title{\LARGE \bf
\textsc{AcroRL}: Learning Aggressive Quadrotor Inversion using Bidirectional Thrust
}

\author{
  Gabriel Rodriguez$^{\ddagger}$, Henri Sayag$^{\ddagger}$, Abhishek
  Rathod$^{\ddagger}$,\\ \textbf{John Stecklein$^{\ddagger}$, Siddharth
  Saha$^{\S}$, Christopher Barngrover$^{\S}$ and Wennie Tabib$^{\ddagger}$} \\
  $^{\ddagger}$Carnegie Mellon University, $^{\S}$Shield AI
}

\begin{document}

\maketitle
\thispagestyle{empty}

\begin{abstract}
Bidirectional thrust grants quadrotors a second equilibrium condition and
increased control authority, expanding the envelope of possible aggressive
maneuvers and enabling inverted flight, perching, and sensing. Prior geometric
control approaches extend differential flatness through Hopf fibration-based
attitude representations to support bidirectional thrust, but struggle with
actuator saturation and motor reversal delay during inversions, requiring
heuristic thrust posture scheduling and waypoint tuning. We propose a
learning-based framework that modulates a constant reference trajectory to
perform \emph{compact}, position-constrained quadrotor  inversions while
remaining compatible with traditional trajectory generation and tracking across
flight regimes. Separate policies are trained via reinforcement learning for
nominal-to-inverted and inverted-to-nominal transitions. In JAX-based
simulation, the proposed method achieves the lowest position deviation and
settling time across all evaluated baselines, reducing position root mean
square error (RMSE) and settling time by at least 25\% and 39\%, respectively,
relative to the strongest optimization-based baseline. Hardware experiments
demonstrate successful inversion across multiple yaw configurations with
position RMSE below \SI{0.33}{\meter}, and compatibility with downstream
trajectory generation and control through circular flight in both regimes.
Additionally, we provide a project website\footnote{Project website available at
\href{https://rislab.github.io/projects/2026-05-14-acro-rl.html}{\nolinkurl{rislab.github.io/projects/2026-05-14-acro-rl.html}}.}
with additional experiment videos and an open-source implementation of the
proposed framework\footnote{Open-source implementation available at
\href{https://github.com/rislab/acrorl}{\nolinkurl{github.com/rislab/acrorl}}.}.

\end{abstract}

\keywords{Reinforcement Learning, Geometric Control, Aerial Robotics}

\section{Introduction}
Traditionally, quadrotors operate around a single equilibrium: upright hover.
Bidirectional electronic speed controllers (ESCs) enable a second equilibrium:
inverted hover. Operating near and transitioning between the two unlocks a
range of desirable capabilities: perching on inclined surfaces, inverted
perching, recovery from large disturbances in either flight regime, and reuse of
a single fixed payload for both ground and overhead tasks, thereby maximizing
low size, weight, and power (SWaP) field efficiency. 

Yet, executing the inversion maneuver remains challenging. Traditional
differentially flat trajectory generation methods, while powerful within either
flight regime, struggle with motor reversal delays introduced by bidirectional
ESCs when switching operating regimes due to control signal jumping
~\citep{yu2020perching, bass2022ral}. As a result, while prior works have 
demonstrated the inversion maneuver, they incur significant position
deviations~\citep{yu2020perching, watterson2020iser, jothiraj2020allocation,  maier2018bidirectional}.

We address this gap in the state of the art by developing aggressive,
positionally \emph{compact} inversions. We introduce the following
contributions: 
i) learned reference modulation policies for aggressive inversion maneuvers
leveraging bidirectional thrust,
ii) a steady-state and randomized transient thrust model for asymmetric
propellers, and 
iii) extensive simulation and hardware evaluation alongside an open-source
release of the proposed approach.

\begin{figure}[t]
    \centering
    \ifthenelse{\equal{\arxivmode}{true}}{}{\vspace{-0.5cm}}
    \includegraphics[width=0.98\textwidth]{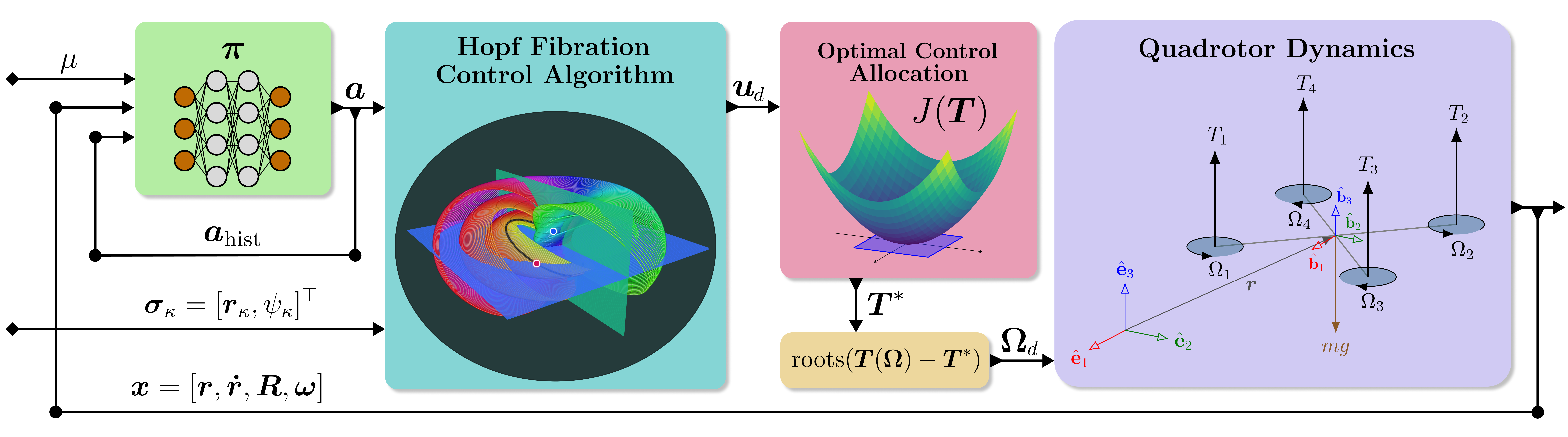}
    \caption{Overview of the proposed method. A reference modulation policy
    $\bm{\pi}$ (Section~\ref{sec:pi}), activated by the inversion flag $\mu$,
    observes the robot state $\bm{x}$ and a finite action history 
    $\bm{a}_\text{hist}$ to produce a position reference modulation and
    continuous thrust posture, $\bm{a} = [\bm{r}_{\delta \kappa},
    \tilde{\eta}]$, whose sign sets the discrete thrust posture $\eta =
    \operatorname{sgn}(\tilde{\eta})$. These outputs and a differentially flat
    reference trajectory $\bm{\sigma}_{\kappa}$ are mapped through a Hopf
    fibration-based control algorithm (Section~\ref{sec:HFCA}) to generate the
    desired control input $\bm{u}_d = [f_c, \bm{\tau}]$. An optimal control
    allocation module (Section~\ref{sec:OCA}) then computes thrust commands
    $\bm{T}^*$ under actuator saturation constraints. Finally, $\bm{T}^*$ is
    converted to desired motor rates $\bm{\Omega}_d$ via an asymmetric thrust
    model and executed on the quadrotor dynamics (Section~\ref{sec:model}).} 
    \label{fig:method}
    \vspace{-10pt} 
\end{figure}

\section{Related Work}
To the best of our knowledge, \textsc{AcroRL} is the first learning-based
approach to address in-place bidirectional thrust quadrotor inversion. This
section provides an overview of recent works in perching and landing, recovery
mechanisms, trajectory generation and control, quadrotor control using
reinforcement learning, and thrust modeling, as these are most aligned with this
work.

\textbf{Perching and Landing:} Recent work has explored leveraging bidirectional
thrust for landing and perching tasks, including landing on inclined
surfaces~\citep{bass2020ral,bass2022ral}, inverted perching on
ceilings~\cite{yu2020perching,yu2022ijmav}, and ceiling effect
perching~\cite{gong2025aggressive}. In contrast to unidirectional thrust  
approaches~\cite{habas2025ceilings,habas2025flies,habas2023invertedlanding,mao2021perch,zou2023perch},
systems that exploit bidirectional thrust exhibit increased control authority
along the body-frame $z$-axis, enabling greater contact forces during surface
interaction as well as increased braking and acceleration capabilities during
approach. However, existing inverted perching maneuvers rely on the vehicle's
approach momentum to complete inversion, resulting in large position deviations.
Such maneuvers are dangerous in constrained and cluttered environments, common
when perching on tree limbs and branches~\cite{li2025design,li2025tail} or near
powerlines~\cite{paneque2022ral}. The proposed approach addresses this
limitation by developing a strategy for \emph{compact}, in-place inversion 
maneuvers to and from inverted hover.

\textbf{Safety and Recovery Mechanisms:}
Recovery after collisions~\cite{battiston2019attitude}, external
disturbances~\cite{mueller2018control}, or throw-and-go
initializations~\cite{faessler2015automatic} is critical for mission execution.
Prior works have leveraged reverse thrust to recover from single-motor
failures~\cite{liao2025ecc}, recover from falls~\cite{zhao2026arxiv}, or improve
quadrotor survivability under external
disturbances~\cite{chen2024icra,zhao2025icca,zhao2026ral}. However, none of
these prior works intentionally utilizes the inverted flight envelope; instead,
reversible thrust is used only as a mechanism for recovery. 

\textbf{Trajectory Generation and Control:}
Existing works utilize a variety of control architectures for bidirectional
thrust quadrotors, including model predictive
control~\cite{wehbeh2022ral,wehbeh2022rsj,wehbeh2024jrnc},
fractional-order PID control~\cite{xu2024science,xu2026jet},
cascaded quaternion-based control~\cite{maier2018bidirectional},
universal UAV control~\cite{jothiraj2019control,jothiraj2020allocation},
and geometric control based on either Gram--Schmidt axis
composition~\cite{yu2020perching,yu2022ijmav} or the Hopf
fibration~\cite{watterson2020iser}. Importantly,~\citet{watterson2020iser}
avoid singularities present in the Gram--Schmidt axis composition via the Hopf
fibration, critical for inversions and downstream tasks. Yet, existing
trajectory generation methods are ill-suited for in-place inversions, relying on
thrust posture heuristics~\cite{mao2023icra}, or optimizing for time while
omitting key transient actuator dynamics~\cite{mao2024toppquad}. In contrast,
our policy explicitly optimizes over the \emph{full} quadrotor and randomized 
actuator dynamics, learning to minimize position deviation directly. 

\textbf{RL-Based Quadrotor Control:} The choice of action space strongly
shapes the optimization landscape and sim2real transfer of RL-based quadrotor
control. Prior works output motor rates~\cite{eschmann2024ral, zhang2023icra,
zhang2025adpatation, han2025ral, eschmann2026science}, single-rotor thrusts
(SRT)~\cite{hwangbo2017ral}, collective thrust and body rates
(CTBR)~\cite{kaufmann2020rss, kaufmann2023nature, heeg2025icra, zhao2026arxiv},
linear velocities (LV)~\cite{belkhale2021ral}, or acceleration/position
setpoints with yaw~\cite{lee2025icra, li2017arxiv}. \citet{kaufmann2022icra}
compare LV, CTBR, and SRT, and find that CTBR yields the highest-performing
policies. Of these works, only~\citet{zhao2026arxiv} consider bidirectional
thrust, pairing CTBR with Incremental Nonlinear Dynamic Inversion (INDI)
tracking. We instead train policies that output position and thrust posture
commands, leaving stabilization to an almost globally stable geometric
controller.  

\textbf{Thrust Modeling:}
Existing works have explicitly utilized symmetric propellers (i.e. 3D or
bidirectional propellers) to guarantee near-symmetric performance across
operating regimes~\citep{jothiraj2019control, jothiraj2020allocation,
ji2025bicopter, watterson2020iser, maier2018bidirectional}, assumed symmetric
performance despite unspecified propeller
geometries~\citep{bass2020ral, zhao2026ral, zhao2025icca, zhao2026arxiv}, or
explicitly captured asymmetry~\citep{chen2024icra, liao2025ecc}. Non-ideal
transient dynamics during motor reversal have been represented using a
deterministic reversal delay~\citep{bass2020ral}, a
dead-zone~\citep{jothiraj2020allocation, jothiraj2019control}, a nonlinear
state-dependent rate constraint~\cite{wehbeh2022ral}, and, more recently, a
stochastic reversal delay~\citep{ji2025bicopter}. While~\citet{ji2025bicopter}
treat motor reversal delay as a stochastic variable, the approach requires
complex log-normal modeling and specialized parameter identification. Similarly,
the deterministic model of~\citet{bass2020ral} relies on genetic optimization
to tune model parameters. In contrast, we propose a model that captures
asymmetric behavior, accounts for stochastic motor reversal delay, and can be
fit using limited experimental data.

\section{Methodology}
\textbf{Notation:}
We adopt the Hamilton convention for quaternions with imaginary basis elements
$(\mathbf{i, j, k})$ satisfying $(\mathbf{ij} = \mathbf{k})$ and represent
quaternions as $\bm{q} = [q_w, q_x, q_y, q_z]^\top$, where $\overline{\bm{q}}$
denotes the quaternion conjugate. Scalars are written in italic typeface ($m$,
$g$), while vectors are represented using lowercase bolded typeface ($\bm{r}$,
$\bm{\omega}$), and matrices using uppercase bolded typeface ($\bm{R}$,
$\bm{P}$). Unit basis vectors of coordinate frames are specifically denoted
with a hat $(\hat{\mathbf{e}}_i, \hat{\mathbf{b}}_i)$. The Hamiltonian
quaternion product is denoted by $\otimes$, $\bm{R}(\bm{q})$ denotes the
rotation matrix associated with $\bm{q}$, and $\bm{R}_x(\cdot)$,
$\bm{R}_y(\cdot)$, $\bm{R}_z(\cdot)$ denote the elementary rotation matrices
about the $x$-, $y$-, and $z$-axes, respectively, such that the ZYX Euler angle
rotation matrix is $\bm{R}(\psi, \theta, \phi) = \bm{R}_z(\psi) \bm{R}_y(\theta)
\bm{R}_x(\phi)$. 

\subsection{System Model}
\label{sec:model}

\textbf{Quadrotor Dynamics:}
We model the dynamics of the quadrotor as a 3D rigid body with control inputs
$\bm{u} = [f_c, \bm{\tau}]$ and state $\bm{x} = [\bm{r}, \bm{\dot{r}}, \bm{R},
\bm{\omega}]$, where $f_c$ is the collective thrust, $\bm{\tau}$ are body
torques, $\bm{r}$ is the position, $\bm{R}$ is the body-to-world rotation
matrix, and $\bm{\omega}$ are body angular velocities. We define an inertial
world frame $\mathcal{W}$ with axes $\hat{\mathbf{e}}_1, \hat{\mathbf{e}}_2,$
and $\hat{\mathbf{e}}_3$, where $\hat{\mathbf{e}}_3$ points opposite the
direction of gravity. The body frame $\mathcal{B}$ is centered at the
quadrotor's center of mass with basis vectors $\hat{\mathbf{b}}_1,
\hat{\mathbf{b}}_2,$ and $\hat{\mathbf{b}}_3$ comprising its axes. The pose of
$\mathcal{B}$ in $\mathcal{W}$ is defined by the translation $\bm{r}$ and
rotation $\bm{R}$. Letting $\bm{\mathcal{I}}$ be the inertia matrix, $g$ the
acceleration due to gravity, and $m$ the mass of the quadrotor, the equations of
motion are given by:
\begin{equation}
    m \bm{\ddot{r}} = -mg \hat{\mathbf{e}}_3 + f_c\hat{\mathbf{b}}_3, \quad\quad
    \bm{\mathcal{I}} \dot{\bm{\omega}} = -\bm{\omega} \times \bm{\mathcal{I}}
    \bm{\omega} + \bm{\tau}. \label{eq:eq_of_motion}
\end{equation}
Individual rotor thrusts $\bm{T}$ are mapped to the control vector $\bm{u}$ by
the relationship $\bm{u} = \bm{M}(\bm{\Omega})\bm{T}$, where
$\bm{M}(\bm{\Omega})$ is the mixer matrix, defined as a function of the motor rate
vector $\bm{\Omega}$:
\begin{equation}
\bm{M}(\bm{\Omega}) =
    \begin{bmatrix}
    1 & 1 & 1 & 1 \\
    y_1 & y_2 & y_3 & y_4 \\
    -\,x_1 & -\,x_2 & -\,x_3 & -\,x_4 \\
      -m_{s_1}(\operatorname{sgn}(\Omega_1)) 
    & -m_{s_2}(\operatorname{sgn}(\Omega_2))
    &  m_{s_3}(\operatorname{sgn}(\Omega_3)) 
    &  m_{s_4}(\operatorname{sgn}(\Omega_4))
    \end{bmatrix}, \label{eq:mixer}
\end{equation}
where $x_i, y_i$ are the positions of the motors in the body frame and $m_{s_i}$
is the ratio of the reaction torque to rotor thrust \cite{mahony2021multirotor},
whose magnitude is dependent on motor spinning direction due to asymmetric rotor
drag characteristics across positive $(+)$ and negative $(-)$ operating regimes
of the propeller.

\textbf{Steady-State Thrust Modeling:}
Executing quadrotor inversions necessitates a thrust model that encompasses both
the $(+)$ and $(-)$ operating regimes of the propeller. Contrary to prior works
limited to the small symmetric propellers available on the market
\citep{jothiraj2019control,jothiraj2020allocation, ji2025bicopter,
watterson2020iser, maier2018bidirectional}, we intentionally utilize larger
asymmetric propellers to maximize peak thrust capability, sacrificing thrust
symmetry in exchange for increased control authority and versatility. Without
sacrificing model fidelity by reducing coefficient number \citep{chen2024icra}
or increasing model order \citep{liao2025ecc}, we propose the following second
order, asymmetric thrust model, where $c_{T_j}$ is the thrust coefficient for
the $j$-th order term \cite{mahony2021multirotor}.
\begin{equation}
    \begin{aligned}
        T_i(\Omega_i) &= c_{T_{2,i}}\Omega_i|\Omega_i| + c_{T_{1,i}}\Omega_i + c_{T_{0,i}} \\
        \tau_{z,i}(\Omega_i) &= m_{s_i} T_i(\Omega_i)
    \end{aligned}
    \quad
    (c_{T_{j,i}}, m_{s_i}) =
    \begin{cases}
        c_{T_{j,i},(+)}, m_{s_i, (+)} & \Omega_i \ge 0 \\
        c_{T_{j,i},(-)}, m_{s_i, (-)} & \Omega_i < 0
    \end{cases}
\end{equation}
Given a desired thrust vector $\bm{T}_d$, the desired motor rate vector
$\bm{\Omega}_d$ is recovered by finding the roots of
$\bm{T}(\bm{\Omega}) - \bm{T}_d$. Since the thrust model is quadratic
in $|\Omega_i|$, we solve for the positive root and restore the sign of the
commanded thrust for each motor:
\begin{equation}
    \Omega_{d,i} = \operatorname{sgn}(T_{d, i})
    \frac{-c_{T_{1,i}} + \sqrt{\max\!\left(c_{T_{1,i}}^2 - 4 c_{T_{2,i}}
    \left(|c_{T_{0,i}}| - |T_{d, i}|\right),\, 0\right)}}
    {2 c_{T_{2,i}}},
\end{equation}
where the $\max(\cdot, 0)$ avoids complex roots, and the coefficients
$c_{T_{j,i}}$ are selected using the motor rates $\bm{\Omega}$ from the previous
time step. In the proposed framework, $\bm{T}_d = \bm{T}^*$ is the
output of the optimal control allocation module (Section~\ref{sec:OCA},
see~\cref{fig:method}). 

\textbf{Transient Thrust Modeling:}
Transient dynamics for asymmetric propellers are also non-trivial and
asymmetric, especially during regime transitions where the dynamics are
dominated by stochastic motor reversal delay (or dead-zone)
\citep{ji2025bicopter, jothiraj2020allocation}. Our objective is not to
explicitly model the exact timing of the reversal delay, but to capture its 
dominant effect on control authority.

Therefore, we adapt the piecewise first-order model presented in
\citep{jothiraj2020allocation} to account for trans-regime transient dynamics
variability, rather than attempting to explicitly model stochastic timing of
reversal. By defining distinct operating regime specific slew rates
$\alpha_{(\pm)}$ and a switching logic $\mathcal{S}$ based on a switching
threshold $\Omega_0$ and dead-zone width $\Delta_\Omega$, we devise a
computationally efficient model that remains structurally simple and easy to fit
to little experimental data. Conditioned on $\mathcal{S}$, the target
equilibrium $\Omega_\text{set}$ is defined as either the true desired motor rate
$\Omega_d$, or the threshold $\Omega_0$ to model the exponential approach to the
dead-zone. 
\begin{align}
    \mathcal{S} &=
        \left(\Omega > \Omega_0 + \Delta_\Omega
            \land \Omega_{d} < \Omega_0 \right) \lor
        \left(\Omega < \Omega_0 - \Delta_\Omega
            \land \Omega_{d} > \Omega_0 \right) \\
    \alpha &=
        \begin{cases}
        \alpha_{(+)}, & \Omega \ge \Omega_0 \\
        \alpha_{(-)}, & \Omega < \Omega_0
        \end{cases}, \quad
    \Omega_{\text{set}} =
        \begin{cases}
        \Omega_0, & \mathcal{S} \\
        \Omega_{d}, & \text{otherwise}
        \end{cases}, \quad
    \dot{\Omega} = \alpha \left( \Omega_{\text{set}} - \Omega \right)
\end{align}
To bridge the sim2real gap, we treat the steady-state and transient thrust model
parameters as random variables sampled from independent uniform distributions
$\mathcal{U}$ during training; namely,
\begin{align}
    \alpha &\sim \mathcal{U}(0.75,\,1.25)\, \alpha_{\text{nom}},
    &c_{T_i} &\sim \mathcal{U}(0.9,\,1.1)\, c_{T_i, \text{nom}}, \\
    \Omega_0 &\sim \mathcal{U}(\Omega_{0,\min},\,\Omega_{0,\max}),
    &\Delta_\Omega &\sim \mathcal{U}(\Delta_{\Omega,\min},\,\Delta_{\Omega,\max}),
\end{align}
where $(\cdot)_\text{nom}$ denotes the experimentally measured parameter. This
domain randomization approach, grounded in the real-world stochasticity of
the transient dynamics (see~\cref{fig:thrust_models}), yields control policies
that are robust to variability in the motor response without needing complex
stochastic modeling \citep{ji2025bicopter} or genetic algorithms for model
fitting \citep{bass2020ral}.

\begin{figure}[b]
\centering
\begin{minipage}{0.48\linewidth}
\centering
\subfloat[Nominal to inverted (NtI) transition]{
\label{sfig:nti_still_frames}
\includegraphics[height=4.7cm]{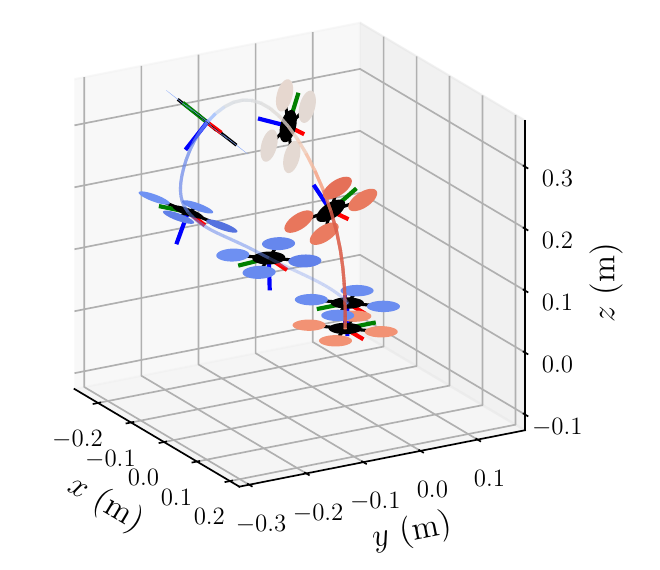}}
\end{minipage}
\hfill
\begin{minipage}{0.48\linewidth}
\centering
\subfloat[Inverted to nominal (ItN) transition]{
\label{sfig:itn_still_frames}
\includegraphics[height=4.7cm]{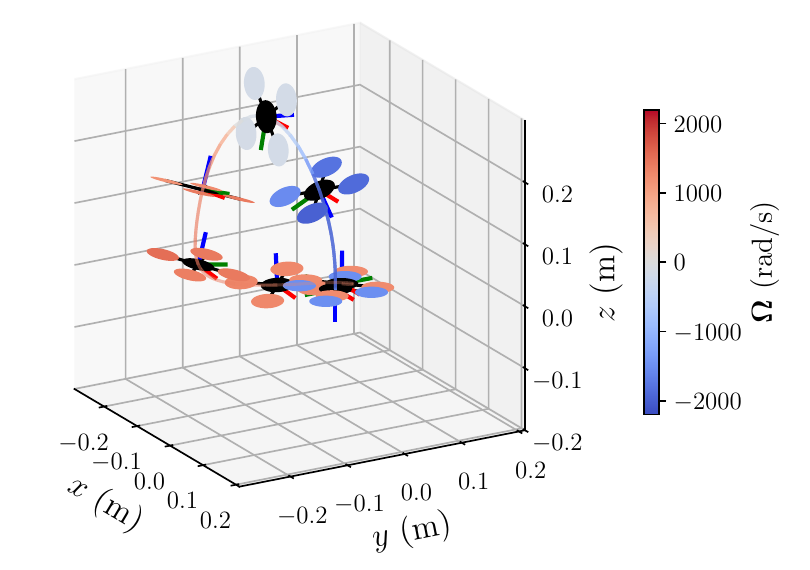}}
\end{minipage}
\caption{\label{fig:still_frames}Still-frame visualizations of the learned
quadrotor inversion trajectories in simulation. The colors encode the rotor
angular rates throughout the trajectory execution (red for positive thrust and 
blue for negative thrust). Compared to
the~\protect\subref{sfig:nti_still_frames} NtI transition, 
the~\protect\subref{sfig:itn_still_frames} ItN transition exhibits a tighter
maneuver profile, which is consistent with the asymmetric thrust profile during
reversal.}  
\end{figure}


\subsection{Learned Inversion Policies (\boldmath{$\pi$})}
\label{sec:pi}

The quadrotor inversion task is formulated as a partially observable Markov
decision process (POMDP), represented by the tuple $\langle\mathcal{O},
\mathcal{A}, \mathcal{P}, \mathcal{R}, \gamma\rangle$. The observation space
$\mathcal{O}$ comprises the quadrotor state $\bm{x}$, and a finite action
history $\bm{a}_\text{hist}$, allowing the policy to infer actuator dynamics and
constraint effects. The action space $\mathcal{A}$ consists of an action $\bm{a} =
[\bm{r}_{\delta\kappa}, \tilde{\eta}]$, where $\bm{r}_{\delta\kappa}$ modulates 
the position component $\bm{r}_\kappa$ of a constant flat reference
$\bm{\sigma}_\kappa = [\bm{r}_\kappa, \psi_\kappa]$, resulting in a position
reference $\tilde{\bm{r}}_\kappa = \bm{r}_\kappa + \bm{r}_{\delta\kappa}$, and
$\tilde{\eta} \in [-1,1]$ is a continuous parametrization of the thrust posture. 
$\mathcal{P}$ denotes the transition probability and $\mathcal{R}: \mathcal{O}
\times \mathcal{A} \rightarrow \mathbb{R}$ is the reward function, with future
rewards discounted by $\gamma$. For training, we define a cost function
$\mathcal{C} = -\mathcal{R}$. $\mathcal{C}$ is designed to produce the
desired positionally \emph{compact} inversion while maintaining smoothness in
control action for favorable sim2real transfer:
\begin{equation}
    \mathcal{C}{=}
    \underbrace{
        w_{\bm{r}} \|\bm{r}\|_1{+}
        w_{\bm{\dot{r}}} L_H(\bm{\dot{r}})}_{\text{Translational Penalties}}{+}
    \underbrace{
        w_{\bm{g}_b} \|\bm{g}_b{-}\bm{g}_{b, d}\|_2{+}
        w_\omega L_H(\bm{\omega})}_{\text{Orientation Penalties}}
    + \underbrace{w_{\tilde{\eta}} \left(1{-}\tilde{\eta}^2\right){+}w_{\bm{\dot{r}}_{\delta \kappa}} \|\bm{r}_{\delta \kappa,t}{-}\bm{r}_{\delta \kappa,t-1}\|_2}_{\text{Action Penalties}},
    \label{eq:reward}
\end{equation}
where $L_H$ is the Huber loss function \citep{huber1992loss}, and $\bm{g}_b =
-\bm{R}^\top \hat{\mathbf{e}}_3$ denotes the gravity vector projected into the
body frame. The orientation penalty is formed using $\bm{g}_b$ to allow the
Hopf fibration-based control algorithm (HFCA) to handle the yaw alignment and
focus the policy's reasoning on inversion. Separate multi-layer perceptron (MLP)
policies $\bm{\pi}_{\bm{\theta}_\text{NtI}}(\bm{a}_t|\bm{x},
\bm{a}_\text{hist})$ and $\bm{\pi}_{\bm{\theta}_\text{ItN}}(\bm{a}_t|\bm{x},
\bm{a}_\text{hist})$, parameterized by $\bm{\theta}_\text{NtI}$ and
$\bm{\theta}_\text{ItN}$, respectively, are trained for the nominal-to-inverted
(NtI) and inverted-to-nominal (ItN) transitions using Proximal Policy
Optimization (PPO) \citep{schulman2017ppo} with separate reward weights
$\bm{w}_\text{NtI}$ and $\bm{w}_\text{ItN}$ in our JAX-based simulator
{\small\texttt{acrorl}} (an adaptation of the simulator presented
in~\citep{heeg2025icra}). Separate policies and weights account for the
asymmetry of the operating regimes, following the use of separate control gains
for different flight regimes in~\citep{yu2020perching}. 

Under bidirectional thrust, \citet{zhao2026arxiv} find that direct motor
commands perform substantially worse than CTBR, and that INDI rate tracking 
outperforms proportional rate tracking through disturbance rejection. For
traditional quadrotors, higher-level action spaces such as LV are less sensitive
to model mismatch than lower-level ones~\citep{kaufmann2022icra}. We therefore
propose an action space $\mathcal{A}$ that modulates the position reference of
the HFCA, which we expect to be less sensitive to model mismatch, aiding
sim2real transfer. Since higher-level action spaces can restrict what the policy
can express (LV policies are outperformed by CTBR~\citep{kaufmann2022icra}), we
set the action limits of $\bm{r}_{\delta\kappa}$ far beyond the controller's
immediate tracking window (see~\cref{sfig:4a}) to encourage
exploration during training. The policy outputs therefore act as
\emph{feedforward} commands that exploit the underlying controller, rather than
as a true reference for direct tracking. Furthermore, including the continuous
thrust posture $\tilde{\eta}$ in $\mathcal{A}$ allows the policy to determine
thrust reversal timing, removing the need for heuristics present in baseline
classical methods.

\subsection{Hopf Fibration-Based Control Algorithm (HFCA)}
\label{sec:HFCA}

\textbf{Position Control:}
The HFCA \citep{watterson2020rr,watterson2020iser} tracks a specified
hybrid-mode differentially flat reference trajectory $\bm{\sigma}_\kappa(t) =
[\bm{r}_\kappa(t), \psi_\kappa(t)]^\top$, conditioned on a thrust posture $\eta \in
\{\pm1\}$, where $\bm{r}_\kappa(t) \in C^3$. We define the translation error
vectors 
\begin{equation}
\bm{e}_{\bm{r}} = \bm{r} - \bm{r}_\kappa, \quad
\bm{e}_{\bm{\dot{r}}} = \bm{\dot{r}} - \bm{\dot{r}}_\kappa, \quad
\bm{e}_{\bm{\ddot{r}}} = \bm{\ddot{r}} - \bm{\ddot{r}}_\kappa,
\end{equation}
when $\bm{\pi}$ is inactive, and redefine both the position error and thrust
posture as $\bm{e}_{\bm{r}} = \bm{r} - \tilde{\bm{r}}_\kappa$ and $\eta =
\operatorname{sgn}(\tilde{\eta})$, respectively,  when
$\bm{\pi}$ is 
activated by the inversion flag $\mu$ (see~\cref{fig:method}). The desired
acceleration $\bm{\ddot{r}}_d$ and jerk $\bm{\dddot{r}}_d$ in the world frame
$\mathcal{W}$ are given by: 
\begin{equation}
    \bm{\ddot{r}}_{d} = - \bm{K}_{\bm{r}} \bm{e}_{\bm{r}} - \bm{K}_{\bm{\dot{r}}} \bm{e}_{\bm{\dot{r}}}
        + \bm{\ddot{r}}_\kappa + g \hat{\bm{e}}_3, \quad \quad
    \bm{\dddot{r}}_{d} = -\bm{K}_{\bm r} \bm{e}_{\bm{\dot{r}}} - \bm{K}_{\bm{\dot{r}}} \bm{e}_{\bm{\ddot{r}}}
        + \bm{\dddot{r}}_\kappa.
\end{equation}
The collective thrust is computed as $f_c =  m\hat{\mathbf{b}}_3 \cdot
\bm{\ddot{r}}_d$. The desired thrust axis $\bm{s}$ and its derivative
$\bm{\dot{s}}$ are:
\begin{equation}
    \bm{s} = \frac{\bm{\ddot{r}}_d}{\|\bm{\ddot{r}}_d\|_2}, \quad \quad
    \bm{\dot{s}} = \frac{1}{\|\bm{\ddot{r}}_d\|_2} \bm{P}\bm{\dddot{r}}_d,
\end{equation}
where $\bm{P} = \bm{I} - \bm{ss}^\top$ is the projection matrix onto the tangent
space of $\Sp^2$. Given the bidirectional thrust capability, $\hat{\mathbf{b}}_3$
is not restricted to be codirectional with $\bm{s}$, only colinear. Thus, we
define the desired body $z$-axis as $\hat{\mathbf{b}}_{3,d} = \eta
\bm{s} = [a, b, c]$, flipping the direction of the desired body $z$-axis in
the case of a commanded negative thrust posture.

\textbf{Attitude Control:}
To cover all of $\text{SO}(3)$ and handle the negative thrust case, we define 4
coordinate charts, as in \citep{watterson2020iser,mao2023icra,zhao2025icca}. The
base quaternions $\bm{q}_{abc, N}$ and $\bm{q}_{abc, S}$ are
\begin{equation}
\bm{q}_{abc,N} = \frac{1}{\sqrt{2(1+c)}} \left[\, 1+c,\ -b,\ a,\ 0 \,\right]^\top, \quad
\bm{q}_{abc,S} = \frac{1}{\sqrt{2(1-c)}} \left[\, -b,\ 1-c,\ 0,\ a \,\right]^\top,
\label{eq:hopf_maps}
\end{equation}
while the yaw quaternion is defined as $\bm{q}_\text{yaw}(\psi) =
[\cos({\psi/2}), 0, 0, \sin({\psi/2})]^\top$, where $\psi \in \{\psi_N,
\psi_S\}$, $\psi_N = \psi_\kappa$, and $\psi_S = 2\operatorname{atan2}(a, b) +
\psi_N$. The desired orientation is then defined as $\bm{q}_d =
\bm{q}_{abc, N} \otimes \bm{q}_\text{yaw}(\psi_N)$ in the first chart, or
$\bm{q}_d = \bm{q}_{abc, S} \otimes \bm{q}_\text{yaw}(\psi_S)$ in the
second chart. The desired angular velocity is then $\bm{\omega}_d = 2
\operatorname{vec} \left( {\bm{q}_d^{-1} \otimes
\bm{\dot{q}}_d} \right).$ The two other charts arise from the case that
$\eta=-1$, where $\psi_N = -\psi_\kappa$ because yaw is now applied about an inverted
axis. To track the desired orientation and angular velocity, we define the
orientation error functions $\bm{e_R} = \operatorname{Log}(\bm{q}_d^{-1}
\otimes \bm{q})$, and $\bm{e_\omega} = \bm{\omega} - \bm{R}(\bm{q}_d^{-1}
\otimes \bm{q})^\top\bm{\omega}_d$, where $\bm{q}$ is the quaternion associated
with $\bm{R}$.

Importantly, diverging from previous works, we use the quaternionic error
function as opposed to the skew-symmetric error function
\citep{mueller2018control} because the latter produces slow convergence when the
error is high \citep{spitzer2020corr}, as during an inversion. Finally, the
orientation feedback is defined as
\begin{equation}
    \bm{\tau} = - \bm{J}_R^{-\top}(\bm{e}_R) \bm{K_R} \bm{e_R} - \bm{K_\omega} \bm{e_\omega},
\end{equation}
where $\bm{J}_R^{-\top}$ is the inverse right-Jacobian implemented as in
\citep{nurlanov2024so3}, defining $\bm{u}_d = [f_c, \bm{\tau}]$ illustrated
in~\cref{fig:method}. We include $\bm{J}_R^{-\top}$ in the feedback law to
account for the curvature of the $\text{SO}(3)$ manifold.

\textbf{Chart Switching Logic:}
For nominal hybrid-mode differentially flat trajectory tracking, we follow the
chart switching logic presented in \citep{watterson2020rr, watterson2020iser},
switching charts at the equator ($c = 0$) using the yaw offset
$2\operatorname{atan2}(a,b)$.

However, step thrust posture commands, such as $c=1$ to $c=-1$ via a
transition in thrust posture sign, generate large instantaneous orientation
error that drives high-speed inversion. Such commands often necessitate chart
transitions at locations where the offset $2\operatorname{atan2}(a,b)$ is
undefined, requiring separate logic to handle this case. Hence, we default to no
offset, which in the $c=1$ to $c=-1$ case, corresponds to $\bm{q}_{abc, S} =
\mathbf{i}$. Since $\mathbf{i} \otimes [0, \cos(-\zeta), \sin(-\zeta), 0]^\top
\otimes \overline{\mathbf{i}} = [0, \cos(\zeta), \sin(\zeta), 0]^\top$, it
follows that no additional yaw offset is introduced, yielding $\psi_S = \psi_N =
-\psi_\kappa$.

\subsection{Optimal Control Allocation (OCA)} \label{sec:OCA}

Without accounting for actuator limits, the desired control input
$\bm{u}_d$ is nominally mapped to desired motor thrusts
$\bm{T}_d$ using direct inversion of the mixer matrix
$\bm{M}(\bm{\Omega})$: $\bm{T}_d = \bm{M}(\bm{\Omega})^{-1}
\bm{u}_d$. \citet{jothiraj2020allocation} cast this control allocation
problem as a weighted least squares formulation with cost
$J(\bm{T}) = \frac{1}{2}\|\bm{W}(\bm{M}(\bm{\Omega})\bm{T} -
\bm{u}_d)\|^2$, where $\bm{W} \in \mathbb{R}^{4 \times 4}$ is a
weighting matrix used to prioritize allocating thrust to roll and pitch body
torques for stabilization under saturation conditions. Using the same relative
weighting, we augment the cost function with a Tikhonov regularization term and
parameter $\lambda$: $J(\bm{T}) = \frac{1}{2}\|\bm{W}(\bm{M}(\bm{\Omega})\bm{T}
- \bm{u}_d)\|^2 + \frac{\lambda}{2}\|\bm{T} - \bm{T}_\text{prev}\|^2$,
where $\bm{T}_\text{prev}$ is the thrust command from the previous time step,
and remove explicit slew-rate constraints as they can impose overly
conservative limits on transient actuator
response~\citep{jothiraj2020allocation}. Expanding the cost and dropping
constant terms, we obtain a similar box-constrained quadratic
objective~\citep{bertsekas2016nonlinear}:
\begin{equation}
\bm{T}^* = \argmin_{\bm{T}} \tfrac12 \bm{T}^\top \bm{H}\bm{T} + \bm{f}^\top
\bm{T} \quad \text{s.t.} \quad \bm{T}_{\min} \le \bm{T} \le \bm{T}_{\max},
\end{equation}
where $\bm{H} = \bm{M}(\bm{\Omega})^\top \bm{W}^\top \bm{W}\bm{M}(\bm{\Omega}) +
\lambda \bm{I}$ and $\bm{f} = -\left(\bm{M}(\bm{\Omega})^\top \bm{W}^\top
\bm{W}\bm{u}_d + \lambda \bm{T}_\text{prev}\right)$. Therefore, when OCA is
used, the desired thrust is $\bm{T}_d = \bm{T}^*$. Contrary
to~\citep{jothiraj2020allocation}, we solve this optimization problem using
Projected Gradient Descent (PGD)~\citep{bertsekas2016nonlinear} for a fixed $k$
iterations to comply with JAX's static-graph compilation methods, enabling
efficient batching during training~\citep{jax2018github}.

\section{Experimental Design and Results}
\begin{table}[t]
\centering
\fontsize{8.5pt}{12pt}\selectfont
\setlength{\tabcolsep}{3pt}
\begin{tabular}{c | l c c c c c c}
\toprule
\makecell{\textbf{inversion} \\ \textbf{type}}
& \textbf{method}
& \makecell{\textbf{success} (\%) \\ {[95\% CI]}}
& \makecell{$\bar{e}_{\bm r}$ (\si{\meter}) \\ {[95\% CI]}}
& \makecell{$t_s$ (\si{\second}) \\ {[95\% CI]}}
& \makecell{$\delta r_x^{\max}\!$ (\si{\meter})}
& \makecell{$\delta r_y^{\max}\!$ (\si{\meter})}
& \makecell{$\delta r_z^{\max}\!$ (\si{\meter})} \\
\midrule

\multirow{6}{*}[0pt]{\shortstack{nominal\\$\downarrow$\\inverted}}
& $S(t){+}$HFCA 
& \makecell{31.8 \\[-3pt] {[27.9,36.0]}}
& \makecell{1.075${\pm}$0.450 \\[-3pt] {[1.005,1.145]}}
& \makecell{\good{2.24${\pm}$0.68} \\[-3pt]  \good{[2.13,2.34]}}
& 1.052
& 1.081
& 1.630 \\  

& $S(t){+}$HFCA${+}$OCA  
& \makecell{\good{53.8} \\[-3pt] \good{[48.4,58.1]}}
& \makecell{1.209${\pm}$0.582 \\[-3pt] {[1.140,1.279]}}
& \makecell{2.27${\pm}$0.69 \\[-3pt] {[2.19,2.35]}}
& 1.134 
& 1.114
& 1.910 \\[-1.5pt]

& $\min \bm{r}^{(4)}{+}$HFCA${+}$OCA 
& \makecell{42.4 \\[-3pt] {[38.1,46.8]}}
& \makecell{\good{0.883${\pm}$0.355} \\[-3pt] \good{[0.835,0.930]}} 
& \makecell{2.87${\pm}$0.61 \\[-3pt] {[2.79,2.95]}}
& \good{0.833}
& \good{0.829} 
& \good{1.374} \\   

& $\bm{\pi}{+}$HFCA${+}$OCA (ours) 
& \makecell{\best{98.6} \\[-3pt] \best{[97.1,99.3]}}
& \makecell{\best{0.488${\pm}$0.158} \\[-3pt] \best{[0.474,0.502]}}
& \makecell{\best{1.29${\pm}$0.25} \\[-3pt] \best{[1.27,1.31]}}
& \best{0.627}
& \best{0.654}
& \best{0.662} \\ 

\midrule

\multirow{4}{*}[0pt]{\shortstack{inverted\\$\downarrow$\\nominal}}
& $S(t){+}$HFCA 
& \makecell{97.0 \\[-3pt] {[95.1,98.1]}}
& \makecell{0.829${\pm}$0.398 \\[-3pt] {[0.793,0.864]}}
& \makecell{1.70${\pm}$0.57 \\[-3pt] {[1.64,1.75]}} 
& 1.054 
& 1.061 
& 1.285 \\  

& $S(t){+}$HFCA${+}$OCA 
& \makecell{\good{97.2} \\[-3pt] \good{[95.3,98.3]}}
& \makecell{0.754${\pm}$0.282 \\[-3pt] {[0.729,0.779]}}
& \makecell{\good{1.58${\pm}$0.52} \\[-3pt] \good{[1.53,1.63]}} 
& 1.021 
& 1.008
& 1.159 \\[-1.5pt]

& $\min \bm{r}^{(4)}{+}$HFCA${+}$OCA 
& \makecell{82.8 \\[-3pt] {[79.2,85.8]}}
& \makecell{\good{0.702${\pm}$0.369} \\[-3pt] \good{[0.666,0.737]}} 
& \makecell{2.23${\pm}$0.53 \\[-3pt] {[2.18,2.28]}}
& \good{0.775} 
& \good{0.778} 
& \good{1.142} \\  

& $\bm{\pi}{+}$HFCA${+}$OCA (ours) 
& \makecell{\best{99.8} \\[-3pt] \best{[98.8,99.9]}}
& \makecell{\best{0.520${\pm}$0.181} \\[-3pt] \best{[0.505,0.536]}} 
& \makecell{\best{1.35${\pm}$0.32} \\[-3pt] \best{[1.32,1.38]}} 
& \best{0.652} 
& \best{0.689} 
& \best{0.674} \\

\bottomrule
\end{tabular}

\vspace{3pt}
\caption{Parallelized method comparison in JAX-based simulation over $n{=}500$
rollouts, where $\bar{e}_{\bm{r}}$ is the position RMSE, $\delta r_x^{\max}$,
$\delta r_y^{\max}$, and $\delta r_z^{\max}$ are the maximum absolute deviations
in $x$-, $y$-, and $z$-position, respectively, and $t_s$ is the $\bm{g}_b$
settling time within a \SI{10}{\degree} cone of $\bm{g}_{b,d}$
(see~\cref{fig:gb_intuition}). Success is the percentage of rollouts that settle
without crashing (95\% Wilson interval). $\bar{e}_{\bm{r}}$ and $t_s$ are mean
$\pm$ standard deviation, with a 95\% confidence interval on the mean in
brackets, while $\delta r^{\max}$ columns are means of per-rollout maxima. Best
results per transition type are in \best{green}, second best in \good{orange}.}
\ifthenelse{\equal{\arxivmode}{true}}
{\vspace{-20pt}}
{\vspace{-10pt}}
\label{tab:simulator_comparison}
\end{table}
\begin{figure}[t]
    \centering
    \includegraphics[width=0.9\textwidth]{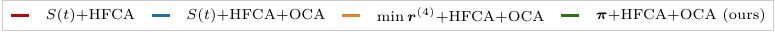}
    \subfloat[Nominal to inverted (NtI) transitions, single rollouts (left), 20 parallel rollouts (right)]{\label{sfig:nti_position_rollouts}\includegraphics[width=0.48\textwidth]{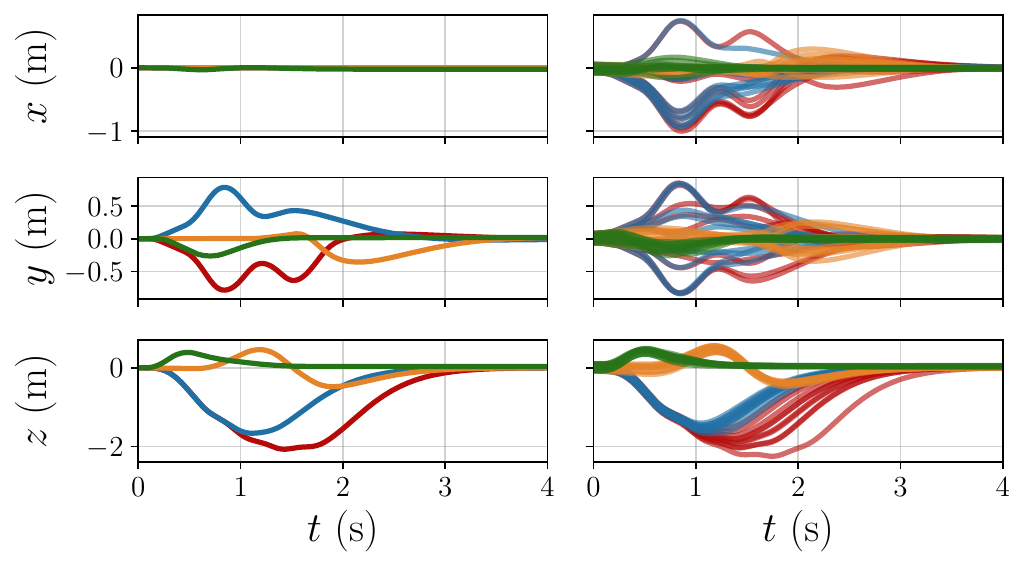}}\hfill %
    \subfloat[Inverted to nominal (ItN) transitions, single rollouts (left), 20 parallel rollouts (right)]{\label{sfig:itn_position_rollouts}\includegraphics[width=0.48\textwidth]{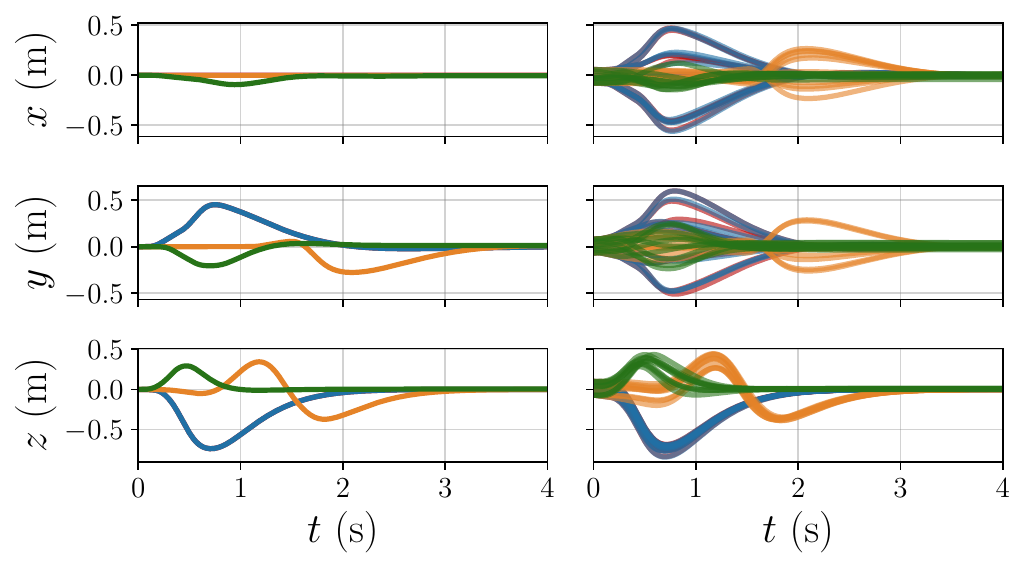}}
    \caption{\label{fig:combined_position_rollouts} Method comparison in
    simulation. Our approach consistently outperforms classical baseline
    inversion strategies for both NtI and ItN transitions in our JAX-based
    simulator {\small\texttt{acrorl}}. Results show single ($n{=}1$) and
    parallel ($n{=}20$) rollout performance across methods. For the sake of
    clarity, single rollouts (left) are shown from ideal hover conditions, while
    parallel rollouts (right) are shown from a random initial state
    distribution. }     
    \vspace{-10pt}
\end{figure}


\textbf{Baseline Inversion Strategies:} \label{sec:baselines}
The proposed method was compared against three baselines: step thrust posture
command with direct inversion control allocation ($S(t)$+HFCA), step thrust
posture with optimal control allocation ($S(t)$+HFCA+OCA), and a minimum-snap
trajectory baseline ($\min \bm{r}^{(4)}{+}$HFCA${+}$OCA). 

The step thrust posture baseline consists of a constant position reference and
a transition from $\eta_\text{init}$ to $-\eta_\text{init}$, inverting
$\hat{\mathbf{b}}_{3,d}$. The minimum-snap baseline follows the method
of~\citep{mao2023icra}, with inversion trajectories defined by three waypoints:
the start position, an increase in $z$ to an apogee height, and a return to the
start position. A change in thrust posture is applied at the second waypoint to
satisfy free-fall constraints. To establish a fair baseline, we conduct a grid
search over the apogee height and the inter-waypoint time intervals, selecting
the parameters that minimize the simulation metrics reported in Table~1. We
restrict the search to three-waypoint trajectories as a practical compromise
between expressivity and tractability.

\textbf{Simulation:} The objective of the simulation experiment is to
characterize relative performance between our method and the baselines for
inversions from both nominal hover $\bm{R}_\text{inv} = \bm{I}$, and inverted
hover $\bm{R}_\text{inv} = \bm{R}_x(\pi)$. The experiment uses $n{=}500$
parallel simulation environments initialized from random initial conditions
according to  
\begin{equation}
    \bm{r}_\text{init} \sim \mathcal{U}(-\Delta\bm{r}, \Delta\bm{r}), \quad
    \bm{\dot{r}}_\text{init} \sim \mathcal{N}(\bm{0}, \sigma_{\dot{r}}^2 \bm{I}), \quad
    \bm{R}_\text{init} \sim \bm{R}(\psi, \theta, \phi) \bm{R}_\text{inv}, \quad
    \bm{\omega}_\text{init} \sim \mathcal{N}(\bm{0}, \sigma_{\omega}^2 \bm{I}),
\end{equation}
where $\Delta\bm{r} = \bm{1} \in \mathbb{R}^3 \si{\meter}$ is the maximum
position perturbation, while $\sigma_{\dot{r}} = \SI{1}{\meter\per\second}$ and
$\sigma_{\omega} = \SI{1}{\radian\per\second}$ are standard deviations in
velocity and body angular rate distributions, respectively. The yaw angle is
sampled according to $\psi \sim \mathcal{U}(-\pi, \pi)$ \si{\radian}, while roll
and  pitch angles are sampled according to $\phi, \theta \sim
\mathcal{U}(-\frac{\pi}{4},\frac{\pi}{4})$ \si{\radian}. 

\begin{wrapfigure}{r}{0.3\textwidth}
    \vspace{-23pt}
    \centering
    \includegraphics[width=0.3\textwidth]{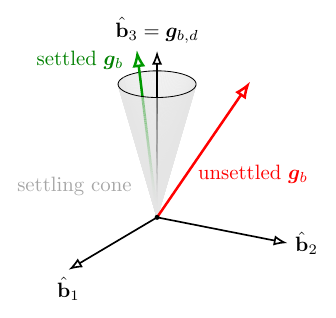}
    \caption{A visualization of the settling cone for an NtI transition.}
    \label{fig:gb_intuition}
    \vspace{-15pt}
\end{wrapfigure}

Single-rollout still frames are illustrated in~\cref{fig:still_frames} while
associated trajectories are shown in~\cref{fig:combined_position_rollouts}. All
quantitative simulation experiment results are reported
in~\cref{tab:simulator_comparison}. We report settling times $t_s$, position
root mean square  error (RMSE) $\bar{e}_{\bm{r}}$, and maximum position
deviations $\max(\delta r_\xi)$ for $\xi \in \{x,y,z\}$. The settling time is
defined as the time required  for the projected gravity vector to enter and
remain within a \SI{10}{\degree}  cone of the target $\bm{g}_{b,d}$ (see
\cref{fig:gb_intuition}), and we define position error metrics relative to the
world origin $\bm{r} = \bm{0} \in \mathbb{R}^3$ for all methods, including the
minimum-snap trajectory baseline. Although counterintuitive for a trajectory
tracking method, this choice is deliberate: the objective is not pure tracking,
but minimal position deviation from the starting hover position.

The learned policy consistently outperforms all three baselines in every
performance metric across all test cases. We attribute this to learning and
optimizing over the true closed-loop system dynamics, specifically the
non-linearities and stochasticity of our transient thrust model. In contrast,
the step command relies on heuristically tuned gains, while the minimum
snap-planner re-parameterizes heuristic tuning to waypoints for a trajectory
that minimizes snap as a proxy for true dynamic feasibility, without considering
actuator constraints. 

Although the step command can produce competitive settling times, it often
results in motor saturation and poor position tracking, as seen in
\cref{sfig:nti_position_rollouts}. Increased performance of the step
command aided by OCA highlights its effectiveness under motor saturation.
It is noted that the policy exhibits small steady-state errors
(\SI{3.52}{\centi\meter} and \SI{4.81}{\centi\meter}) for $n{=}20$ ItN 
and NtI transitions, respectively, which we attribute to
competing terms in the reward function, a noted limitation of RL-based quadrotor
control \cite{lopes2018ppo}.

\begin{figure}[t]
    \centering
    \begin{minipage}[t]{0.45\textwidth}
      \vspace{0cm} 
        \centering
        \begin{tikzpicture}
            \matrix (legend) [draw=black!20, fill=white, inner sep=0.5ex, column sep=1.7ex,
                row sep=1.0ex, nodes={font=\scriptsize}] {
                \draw[yawA, line width=1.5pt] (0,0) -- (0.5,0); & \node[inner sep=0pt] {$\psi \approx 0$}; &
                \draw[yawB, line width=1.5pt] (0,0) -- (0.5,0); & \node[inner sep=0pt] {$\psi \approx -\pi/4$}; &
                \draw[yawC, line width=1.5pt] (0,0) -- (0.5,0); & \node[inner sep=0pt] {$\psi \approx \pi$}; \\
            };
        \end{tikzpicture}
        \subfloat[Hardware inversions at varying yaw angles for
        NtI (left) and ItN (right) transitions. Solid: measured position;
        dashed: commanded position by $\bm{\pi}$.]{\label{sfig:4a}\includegraphics[height=5.35cm]{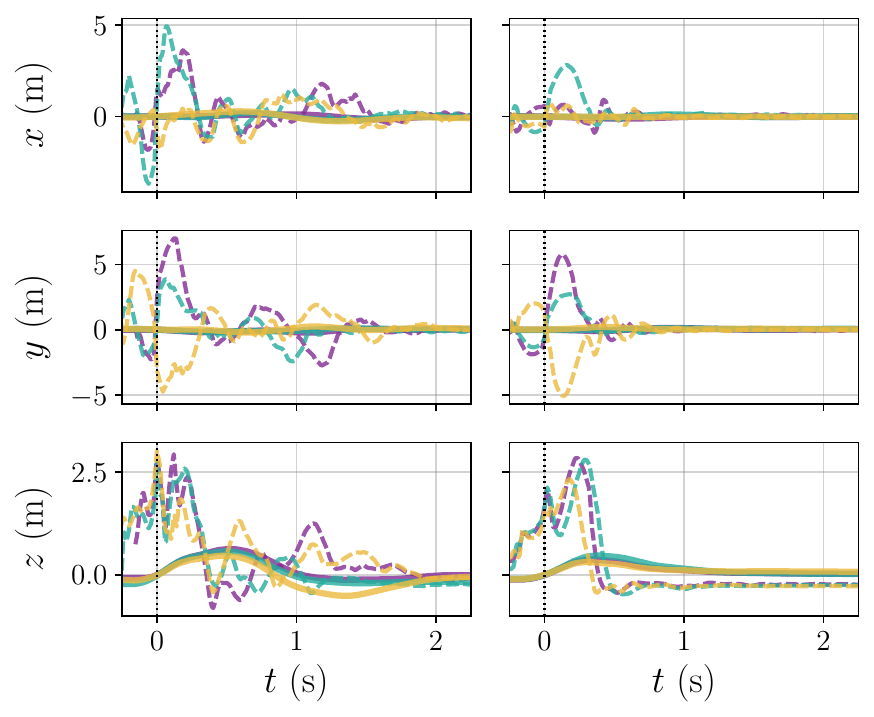}}
    \end{minipage} \hfill%
    \begin{minipage}[t]{0.53\textwidth}
      \vspace{17.75pt}
        \centering
        \subfloat[Circular trajectories after a NtI transition (top)
        and an ItN transition (bottom). Colors encode rotor angular
        rates along the trajectory.]{%
          \label{sfig:4b}
        \begin{minipage}[c]{0.73\linewidth}
            \includegraphics[height=2.75cm]{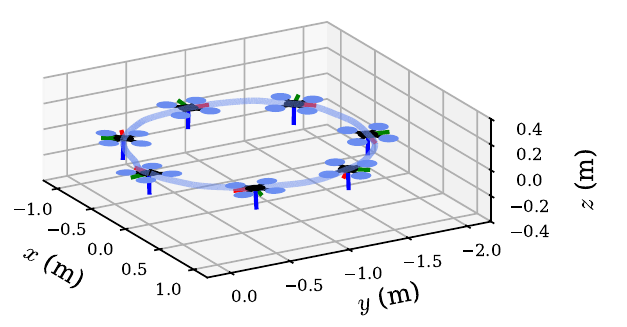}\\
            \includegraphics[height=2.75cm]{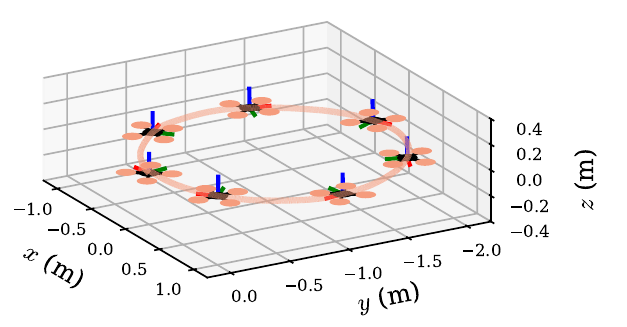}
        \end{minipage}%
        \hspace{0.005\linewidth}%
        \begin{minipage}[c]{0.18\linewidth}
            \includegraphics[height=3.85cm]{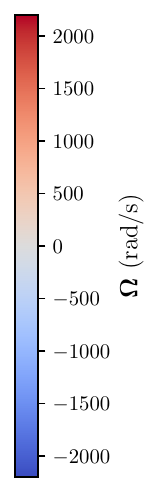}
        \end{minipage}%
        }
    \end{minipage}%
    \caption{Hardware experiments demonstrating the proposed method.~\protect\subref{sfig:4a}
      Inversion trajectories at three initial yaw angles ($n{=}3$ trials per
      transition type) demonstrate that the proposed method can reliably invert
      at varying yaw angles, as required for following general trajectories
      before and after transition.~\protect\subref{sfig:4b} Circular
      trajectories flown after each transition type demonstrate that the
      proposed method is compatible with the HFCA.} 
    \label{fig:hardware_results}
    \ifthenelse{\equal{\arxivmode}{true}}
{}
{\vspace{-10pt}}
\end{figure}

\textbf{Hardware:} Two hardware experiments were performed to validate the
proposed method. Baselines were evaluated only in simulation, as they exhibited
instability and deploying them risked damage to the platform. The first tested
the NtI and ItN transitions at varying yaw angles, evaluating position deviation
during inversion while maintaining yaw using the HFCA. The second validated the
full pipeline by performing yaw-tracking circular trajectories before and after
inversions. Results from both experiments are visualized
in~\cref{fig:hardware_results}, and quantitatively summarized
in~\cref{tab:hardware_results}. Since hardware flips are not initialized at the
true world origin, position error metrics are computed relative to the inversion
starting point $\bm{r}_\text{init}$ over the same duration as the simulation
rollouts. 

\begin{table}[b]
\centering
\footnotesize
\setlength{\tabcolsep}{10pt}
\begin{tabular}{c | l c c c c c}
\toprule 
\textbf{inversion type} & \textbf{initial yaw} & $\bar{e}_{\bm r}$ (\si{\meter})
& $t_s$ (\si{\second}) & $\delta r_x^{\max}\!$ (\si{\meter}) & $\delta
r_y^{\max}\!$ (\si{\meter}) & $\delta r_z^{\max}\!$ (\si{\meter}) \\ 
\midrule
\multirow{3}{*}[0pt]{\shortstack{nominal\\$\downarrow$\\inverted}}
& $\psi \approx 0$      & \best{0.2557} & \best{2.2040} & \best{0.1501} & \best{0.2430} & \good{0.7074} \\ 
& $\psi \approx -\pi/4$ & 0.3233        & 2.8566        & \good{0.2872} & \good{0.2577} & 0.7309        \\
& $\psi \approx \pi$    & \good{0.2737} & \good{2.3222} & 0.3498        & 0.2975        & \best{0.5834} \\
\midrule
\multirow{3}{*}[0pt]{\shortstack{inverted\\$\downarrow$\\nominal}}
& $\psi \approx 0$      & \best{0.2236} & 1.3681        & 0.1320        & \good{0.1389} & \good{0.4924} \\
& $\psi \approx -\pi/4$ & \good{0.2431} & \good{1.2754} & \good{0.1080} & \best{0.1267} & 0.5766        \\
& $\psi \approx \pi$    & 0.2563        & \best{1.2353} & \best{0.0823} & 0.2090        & \best{0.4200} \\ 
\bottomrule
\end{tabular}
\vspace{0.2em}
\caption{Hardware inversion results for the proposed method across three initial
yaw configurations (one trial per configuration), where $\bar{e}_{\bm{r}}$ is the position RMSE, $\delta r_x^{\max}$,
$\delta r_y^{\max}$, and $\delta r_z^{\max}$ are the maximum absolute
deviations in $x$-, $y$-, and $z$-position, respectively, and $t_s$ is the $\bm{g}_b$ settling time within a
\SI{10}{\degree} cone of $\bm{g}_{b,d}$. Best results per transition type are
highlighted in \best{green}, second best in \good{orange}. }
\label{tab:hardware_results} 
\end{table}

The proposed method transfers zero-shot to hardware, achieving successful
inversions with position RMSE below \SI{0.33}{\meter} across all initial yaw
configurations. Given that the simulation initial conditions are intentionally
broader than the near-hover hardware starting states, absolute errors are not
directly comparable between the two settings; thus, we instead compare the
relative behavior of the two transition types. In simulation, NtI and ItN
performance is roughly symmetric. On hardware, however, NtI is consistently
worse: mean settling time is 91\% longer than ItN (\SI{2.46}{\second} vs.\
\SI{1.29} {\second}), and $\delta r_x^{\max}$, $\delta r_y^{\max}$, and $\delta
r_z^{\max}$ are roughly 2.4$\times$, 1.7$\times$, and 1.4$\times$ larger,
respectively. This asymmetry is consistent with adverse propeller geometry in
the inverted regime inducing rotor-body-rotor aerodynamic interactions that
degrade performance and are absent from the simulation environments.

For the circular trajectory experiment, the proposed HFCA chart-switching logic
enables world-frame yaw tracking throughout the trajectory, both
before and after inversion transitions, demonstrating that the full pipeline is
suitable for agile trajectory following across flight regimes.


\section{Conclusion}
We presented AcroRL, a learning-based method to perform \emph{compact} quadrotor
inversions using bidirectional thrust. By training separate reference modulation
policies using PPO, we achieve in-place inversions while remaining compatible with
traditional geometric control techniques for executing arbitrary trajectories
before and after inversion. Careful modeling of asymmetric steady-state thrust
characteristics and stochastic reversal delay during training yields favorable
sim2real transfer without requiring complex stochastic modeling or genetic
optimization. 

In simulation, our method consistently outperforms baseline classical methods in
position RMSE, maximum position deviation, and settling times to inversion. More
specifically, our method reduces RMSE by at least 25\% and settling
time by 39\% relative to a minimum snap optimization-based
baseline. On hardware, we perform successful inversions at varying yaw angles
with position RMSE below \SI{0.33}{\meter}, and demonstrate our framework's
ability to perform a yaw-tracking circular trajectory in both flight regimes,
before and after inversion.

\textbf{Limitations:}
A key limitation of the approach is the observed sim2real gap, specifically
concerning unmodeled aerodynamic effects during inverted-regime asymmetric
propeller operation. Additionally, the absence of an integrator in the reference
controller produces steady-state position offsets on hardware. Finally, the
current reward structure is tailored to inversion maneuvers, and extending to
more general acrobatic behaviors would require reward reshaping and retuning.

\ifthenelse{\equal{\arxivmode}{true}}
{}
{\clearpage}

\acknowledgments{This work was supported in part by the Army Research
Laboratory and the Army Research Office under contract/grant number
W911NF-25-2-0153, AI2C Seed grant, and gift funding from Shield AI.}

\bibliography{refs}

\begin{thebibliography}{60}
\providecommand{\natexlab}[1]{#1}
\providecommand{\url}[1]{\texttt{#1}}
\expandafter\ifx\csname urlstyle\endcsname\relax
  \providecommand{\doi}[1]{doi: #1}\else
  \providecommand{\doi}{doi: \begingroup \urlstyle{rm}\Url}\fi

\bibitem[Yu et~al.(2020)Yu, Chamitoff, and Wong]{yu2020perching}
P.~Yu, G.~Chamitoff, and K.~Wong.
\newblock Perching upside down with bi-directional thrust quadrotor.
\newblock In \emph{2020 International Conference on Unmanned Aircraft Systems
  (ICUAS)}, pages 1697--1703, 2020.
\newblock \doi{10.1109/ICUAS48674.2020.9213946}.

\bibitem[Bass et~al.(2022)Bass, Tunney, and Desbiens]{bass2022ral}
J.~Bass, I.~Tunney, and A.~L. Desbiens.
\newblock Adaptative friction shock absorbers and reverse thrust for fast
  multirotor landing on inclined surfaces.
\newblock \emph{IEEE Robotics and Automation Letters}, 7\penalty0 (3):\penalty0
  6701--6708, 2022.
\newblock \doi{10.1109/LRA.2022.3176102}.

\bibitem[Watterson et~al.(2020)Watterson, Zahra, and Kumar]{watterson2020iser}
M.~Watterson, A.~Zahra, and V.~Kumar.
\newblock Geometric control and trajectory optimization for bidirectional
  thrust quadrotors.
\newblock In J.~Xiao, T.~Kr{\"o}ger, and O.~Khatib, editors, \emph{Proceedings
  of the 2018 International Symposium on Experimental Robotics}, pages
  165--176, Cham, 2020. Springer International Publishing.
\newblock ISBN 978-3-030-33950-0.

\bibitem[Jothiraj et~al.(2020)Jothiraj, Sharf, and
  Nahon]{jothiraj2020allocation}
W.~Jothiraj, I.~Sharf, and M.~Nahon.
\newblock Control allocation of bidirectional thrust quadrotor subject to
  actuator constraints.
\newblock In \emph{2020 International Conference on Unmanned Aircraft Systems
  (ICUAS)}, pages 932--938, 2020.
\newblock \doi{10.1109/ICUAS48674.2020.9214036}.

\bibitem[Maier(2018)]{maier2018bidirectional}
M.~Maier.
\newblock Bidirectional thrust for multirotor mavs with fixed-pitch propellers.
\newblock In \emph{2018 IEEE/RSJ International Conference on Intelligent Robots
  and Systems (IROS)}, pages 1--8, 2018.
\newblock \doi{10.1109/IROS.2018.8593836}.

\bibitem[Bass and Desbiens(2020)]{bass2020ral}
J.~Bass and A.~L. Desbiens.
\newblock Improving multirotor landing performance on inclined surfaces using
  reverse thrust.
\newblock \emph{IEEE Robotics and Automation Letters}, 5\penalty0 (4):\penalty0
  5850--5857, 2020.
\newblock \doi{10.1109/LRA.2020.3010208}.

\bibitem[Yu and Wong(2022)]{yu2022ijmav}
P.~Yu and K.~Wong.
\newblock An implementation framework for vision-based bat-like inverted
  perching with bi-directional thrust quadrotor.
\newblock \emph{International Journal of Micro Air Vehicles}, 14:\penalty0
  1--12, 2022.
\newblock \doi{10.1177/17568293211073672}.
\newblock URL \url{https://doi.org/10.1177/17568293211073672}.

\bibitem[Gong et~al.(2025)Gong, Shao, Li, Wang, and Wang]{gong2025aggressive}
M.~Gong, Z.~Shao, B.~Li, J.~Wang, and Y.~Wang.
\newblock Aggressive perching trajectory planning and control for quadrotor.
\newblock \emph{IFAC-PapersOnLine}, 59\penalty0 (20):\penalty0 1243--1248,
  2025.

\bibitem[Habas et~al.(2025)Habas, Brown, Lee, Goldman, and
  Cheng]{habas2025ceilings}
B.~Habas, A.~Brown, D.~Lee, M.~Goldman, and B.~Cheng.
\newblock From ceilings to walls: Universal dynamic perching of quadrotors on
  surfaces with variable orientations.
\newblock In \emph{2025 IEEE International Conference on Robotics and
  Automation (ICRA)}, pages 288--294, 2025.
\newblock \doi{10.1109/ICRA55743.2025.11128577}.

\bibitem[Habas and Cheng(2025)]{habas2025flies}
B.~Habas and B.~Cheng.
\newblock From flies to robots: Inverted landing in small quadcopters with
  dynamic perching.
\newblock \emph{IEEE Transactions on Robotics}, 41:\penalty0 1773--1790, 2025.
\newblock \doi{10.1109/TRO.2025.3543263}.

\bibitem[Habas et~al.(2023)Habas, Langelaan, and
  Cheng]{habas2023invertedlanding}
B.~Habas, J.~W. Langelaan, and B.~Cheng.
\newblock Inverted landing in a small aerial robot via deep reinforcement
  learning for triggering and control of rotational maneuvers.
\newblock In \emph{2023 IEEE International Conference on Robotics and
  Automation (ICRA)}, pages 3368--3375, 2023.
\newblock \doi{10.1109/ICRA48891.2023.10160376}.

\bibitem[Mao et~al.(2021)Mao, Li, Nogar, Kroninger, and Loianno]{mao2021perch}
J.~Mao, G.~Li, S.~Nogar, C.~Kroninger, and G.~Loianno.
\newblock Aggressive visual perching with quadrotors on inclined surfaces.
\newblock In \emph{2021 IEEE/RSJ International Conference on Intelligent Robots
  and Systems (IROS)}, pages 5242--5248, 2021.
\newblock \doi{10.1109/IROS51168.2021.9636690}.

\bibitem[Zou et~al.(2023)Zou, Li, Ren, Xu, Li, Cai, Zhou, and
  Zhang]{zou2023perch}
Y.~Zou, H.~Li, Y.~Ren, W.~Xu, Y.~Li, Y.~Cai, S.~Zhou, and F.~Zhang.
\newblock Perch a quadrotor on planes by the ceiling effect.
\newblock In \emph{2023 IEEE 19th International Conference on Automation
  Science and Engineering (CASE)}, pages 1--7, 2023.
\newblock \doi{10.1109/CASE56687.2023.10260542}.

\bibitem[Li et~al.(2025{\natexlab{a}})Li, Liu, Zhu, Zhang, Luo, Wang, Liu, and
  Zhao]{li2025design}
Y.~Li, D.~Liu, Y.~Zhu, J.~Zhang, Y.~Luo, Z.~Wang, C.~Liu, and J.~Zhao.
\newblock Design and control of a perching drone inspired by the prey-capturing
  mechanism of venus flytrap.
\newblock \emph{arXiv preprint arXiv:2509.13249}, 2025{\natexlab{a}}.

\bibitem[Li et~al.(2025{\natexlab{b}})Li, Li, Zhang, Huang, Wang, and
  Cai]{li2025tail}
Y.~Li, D.~Li, Y.~Zhang, L.~Huang, S.~Wang, and B.~Cai.
\newblock Tail-mav: Design and control of a perching micro aerial vehicle
  inspired by the tail-suspended behavior of primates.
\newblock In \emph{2025 7th International Symposium on Robotics and Intelligent
  Manufacturing Technology (ISRIMT)}, pages 372--376, 2025{\natexlab{b}}.
\newblock \doi{10.1109/ISRIMT67769.2025.11413171}.

\bibitem[Paneque et~al.(2022)Paneque, Dios, Ollero, Hanover, Sun, Romero, and
  Scaramuzza]{paneque2022ral}
J.~L. Paneque, J.~R. M.-d. Dios, A.~Ollero, D.~Hanover, S.~Sun, A.~Romero, and
  D.~Scaramuzza.
\newblock Perception-aware perching on powerlines with multirotors.
\newblock \emph{IEEE Robotics and Automation Letters}, 7\penalty0 (2):\penalty0
  3077--3084, 2022.
\newblock \doi{10.1109/LRA.2022.3145514}.

\bibitem[Battiston et~al.(2019)Battiston, Sharf, and
  Nahon]{battiston2019attitude}
A.~Battiston, I.~Sharf, and M.~Nahon.
\newblock Attitude estimation for collision recovery of a quadcopter unmanned
  aerial vehicle.
\newblock \emph{The International Journal of Robotics Research}, 38\penalty0
  (10-11):\penalty0 1286--1306, 2019.

\bibitem[Mueller(2018)]{mueller2018control}
M.~W. Mueller.
\newblock Multicopter attitude control for recovery from large disturbances,
  2018.
\newblock URL \url{https://arxiv.org/abs/1802.09143}.

\bibitem[Faessler et~al.(2015)Faessler, Fontana, Forster, and
  Scaramuzza]{faessler2015automatic}
M.~Faessler, F.~Fontana, C.~Forster, and D.~Scaramuzza.
\newblock Automatic re-initialization and failure recovery for aggressive
  flight with a monocular vision-based quadrotor.
\newblock In \emph{2015 IEEE International Conference on Robotics and
  Automation (ICRA)}, pages 1722--1729, 2015.
\newblock \doi{10.1109/ICRA.2015.7139420}.

\bibitem[Liao et~al.(2025)Liao, Neo, Peng, Jia, Yash, and Liu]{liao2025ecc}
F.~Liao, D.~Neo, K.~Peng, D.~Jia, A.~Yash, and W.~Liu.
\newblock Reversible thrust-based fault tolerant control for quadrotor uavs
  against motor failure.
\newblock In \emph{2025 European Control Conference (ECC)}, pages 1531--1536,
  2025.
\newblock \doi{10.23919/ECC65951.2025.11187187}.

\bibitem[Zhao et~al.(2026)Zhao, Zhong, Hoi, Mou, Wang, Wang, Hou, and
  Gao]{zhao2026arxiv}
A.~Zhao, Y.~Zhong, K.~Hoi, J.~Mou, J.~Wang, L.~Wang, J.~Hou, and F.~Gao.
\newblock Agile fall recovery for quadrotors with bidirectional thrust via
  reinforcement learning, 2026.
\newblock URL \url{https://arxiv.org/abs/2606.16513}.

\bibitem[Chen et~al.(2024)Chen, Mo, Zhang, Li, and Cheng]{chen2024icra}
Z.~Chen, S.~Mo, B.~Zhang, J.~Li, and H.~Cheng.
\newblock Robust control for bidirectional thrust quadrotors under
  instantaneously drastic disturbances.
\newblock In \emph{2024 IEEE International Conference on Robotics and
  Automation (ICRA)}, pages 6186--6192, 2024.
\newblock \doi{10.1109/ICRA57147.2024.10611241}.

\bibitem[Zhao et~al.(2025)Zhao, Lyu, and Huang]{zhao2025icca}
Y.~Zhao, M.~Lyu, and H.~Huang.
\newblock A novel anti-disturbance control framework for bidirectional
  quadrotors.
\newblock In \emph{2025 IEEE 19th International Conference on Control \&
  Automation (ICCA)}, pages 310--315, 2025.
\newblock \doi{10.1109/ICCA65672.2025.11129814}.

\bibitem[Zhao et~al.(2026)Zhao, Lyu, Li, and Huang]{zhao2026ral}
Y.~Zhao, M.~Lyu, C.~Li, and H.~Huang.
\newblock Bidirectional thrust control for quadrotor safety.
\newblock \emph{IEEE Robotics and Automation Letters}, 11\penalty0
  (3):\penalty0 2650--2657, 2026.
\newblock \doi{10.1109/LRA.2026.3653327}.

\bibitem[Wehbeh and Sharf(2022{\natexlab{a}})]{wehbeh2022ral}
J.~Wehbeh and I.~Sharf.
\newblock An mpc formulation on $so(3)$ for a quadrotor with bidirectional
  thrust and nonlinear thrust constraints.
\newblock \emph{IEEE Robotics and Automation Letters}, 7\penalty0 (2):\penalty0
  4945--4952, 2022{\natexlab{a}}.
\newblock \doi{10.1109/LRA.2022.3154021}.

\bibitem[Wehbeh and Sharf(2022{\natexlab{b}})]{wehbeh2022rsj}
J.~Wehbeh and I.~Sharf.
\newblock Geometric mpc techniques for reduced attitude control on quadrotors
  with bidirectional thrust.
\newblock In \emph{2022 IEEE/RSJ International Conference on Intelligent Robots
  and Systems (IROS)}, pages 12330--12335, 2022{\natexlab{b}}.
\newblock \doi{10.1109/IROS47612.2022.9982250}.

\bibitem[Wehbeh and Sharf(2024)]{wehbeh2024jrnc}
J.~Wehbeh and I.~Sharf.
\newblock Nonlinear scenario-based model predictive control for quadrotors with
  bidirectional thrust.
\newblock \emph{International Journal of Robust and Nonlinear Control},
  34\penalty0 (18):\penalty0 12450--12475, 2024.
\newblock \doi{https://doi.org/10.1002/rnc.7627}.
\newblock URL \url{https://onlinelibrary.wiley.com/doi/abs/10.1002/rnc.7627}.

\bibitem[Xu et~al.(2024)Xu, Cai, Wang, and Shen]{xu2024science}
L.~Xu, Z.~Cai, Y.~Wang, and Z.~Shen.
\newblock The control method of a quadrotor driven by bidirectional electronic
  speed controllers.
\newblock \emph{Scientific Reports}, 14\penalty0 (1):\penalty0 19532, 2024.
\newblock \doi{10.1038/s41598-024-70681-3}.
\newblock URL \url{https://doi.org/10.1038/s41598-024-70681-3}.

\bibitem[Xu et~al.(2026)Xu, Cai, Wang, Cai, and Liu]{xu2026jet}
L.~Xu, Z.~Cai, Y.~Wang, R.~Cai, and Y.~Liu.
\newblock The motion planning, learning and control of a bidirectional thrust
  quadrotor with special tasks.
\newblock \emph{Journal of Electrical Engineering \& Technology}, 21\penalty0
  (3):\penalty0 3043--3059, 2026.
\newblock \doi{10.1007/s42835-026-02591-5}.
\newblock URL \url{https://doi.org/10.1007/s42835-026-02591-5}.

\bibitem[Jothiraj et~al.(2019)Jothiraj, Miles, Bulka, Sharf, and
  Nahon]{jothiraj2019control}
W.~Jothiraj, C.~Miles, E.~Bulka, I.~Sharf, and M.~Nahon.
\newblock Enabling bidirectional thrust for aggressive and inverted quadrotor
  flight.
\newblock In \emph{2019 International Conference on Unmanned Aircraft Systems
  (ICUAS)}, pages 534--541, 2019.
\newblock \doi{10.1109/ICUAS.2019.8798234}.

\bibitem[Mao et~al.(2023)Mao, Welde, Hsieh, and Kumar]{mao2023icra}
K.~Mao, J.~Welde, M.~A. Hsieh, and V.~Kumar.
\newblock Trajectory planning for the bidirectional quadrotor as a
  differentially flat hybrid system.
\newblock In \emph{2023 IEEE International Conference on Robotics and
  Automation (ICRA)}, pages 1242--1248, 2023.
\newblock \doi{10.1109/ICRA48891.2023.10160320}.

\bibitem[Mao et~al.(2024)Mao, Spasojevic, Hsieh, and Kumar]{mao2024toppquad}
K.~Mao, I.~Spasojevic, M.~A. Hsieh, and V.~Kumar.
\newblock Toppquad: Dynamically-feasible time-optimal path parametrization for
  quadrotors.
\newblock In \emph{2024 IEEE/RSJ International Conference on Intelligent Robots
  and Systems (IROS)}, pages 13136--13143. IEEE, 2024.

\bibitem[Eschmann et~al.(2024)Eschmann, Albani, and Loianno]{eschmann2024ral}
J.~Eschmann, D.~Albani, and G.~Loianno.
\newblock Learning to fly in seconds.
\newblock \emph{IEEE Robotics and Automation Letters}, pages 1--8, 2024.
\newblock \doi{10.1109/LRA.2024.3396025}.

\bibitem[Zhang et~al.(2023)Zhang, Loquercio, Wu, Kumar, Malik, and
  Mueller]{zhang2023icra}
D.~Zhang, A.~Loquercio, X.~Wu, A.~Kumar, J.~Malik, and M.~W. Mueller.
\newblock Learning a single near-hover position controller for vastly different
  quadcopters.
\newblock In \emph{2023 IEEE International Conference on Robotics and
  Automation (ICRA)}, pages 1263--1269, 2023.
\newblock \doi{10.1109/ICRA48891.2023.10160836}.

\bibitem[Zhang et~al.(2025)Zhang, Loquercio, Tang, Wang, Malik, and
  Mueller]{zhang2025adpatation}
D.~Zhang, A.~Loquercio, J.~Tang, T.-H. Wang, J.~Malik, and M.~W. Mueller.
\newblock A learning-based quadcopter controller with extreme adaptation.
\newblock \emph{IEEE Transactions on Robotics}, 41:\penalty0 3948--3964, 2025.
\newblock \doi{10.1109/TRO.2025.3577037}.

\bibitem[Han et~al.(2025)Han, Huang, Xu, Zhang, Yuze, Wang, Wu, and
  Gao]{han2025ral}
Z.~Han, X.~Huang, Z.~Xu, J.~Zhang, W.~Yuze, M.~Wang, T.~Wu, and F.~Gao.
\newblock Reactive aerobatic flight via reinforcement learning.
\newblock \emph{IEEE Robotics and Automation Letters}, 10:\penalty0
  11014--11021, 01 2025.
\newblock \doi{10.1109/LRA.2025.3606383}.

\bibitem[Eschmann et~al.(2026)Eschmann, Albani, and
  Loianno]{eschmann2026science}
J.~Eschmann, D.~Albani, and G.~Loianno.
\newblock Raptor: A foundation policy for quadrotor control.
\newblock \emph{Science Robotics}, 11\penalty0 (114):\penalty0 eaec1481, 2026.
\newblock \doi{10.1126/scirobotics.aec1481}.
\newblock URL
  \url{https://www.science.org/doi/abs/10.1126/scirobotics.aec1481}.

\bibitem[Hwangbo et~al.(2017)Hwangbo, Sa, Siegwart, and Hutter]{hwangbo2017ral}
J.~Hwangbo, I.~Sa, R.~Siegwart, and M.~Hutter.
\newblock Control of a quadrotor with reinforcement learning.
\newblock \emph{IEEE Robotics and Automation Letters}, 2\penalty0 (4):\penalty0
  2096–2103, Oct. 2017.
\newblock ISSN 2377-3774.
\newblock \doi{10.1109/lra.2017.2720851}.
\newblock URL \url{http://dx.doi.org/10.1109/LRA.2017.2720851}.

\bibitem[Kaufmann et~al.(2020)Kaufmann, Loquercio, Ranftl, M{\"u}ller, Koltun,
  and Scaramuzza]{kaufmann2020rss}
E.~Kaufmann, A.~Loquercio, R.~Ranftl, M.~M{\"u}ller, V.~Koltun, and
  D.~Scaramuzza.
\newblock Deep drone acrobatics.
\newblock In \emph{Proceedings of Robotics: Science and Systems}, Corvalis,
  Oregon, USA, July 2020.
\newblock \doi{10.15607/RSS.2020.XVI.040}.

\bibitem[Kaufmann et~al.(2023)Kaufmann, Bauersfeld, Loquercio, Mueller, Koltun,
  and Scaramuzza]{kaufmann2023nature}
E.~Kaufmann, L.~Bauersfeld, A.~Loquercio, M.~Mueller, V.~Koltun, and
  D.~Scaramuzza.
\newblock Champion-level drone racing using deep reinforcement learning.
\newblock \emph{Nature}, 620:\penalty0 982--987, 08 2023.
\newblock \doi{10.1038/s41586-023-06419-4}.

\bibitem[Heeg et~al.(2025)Heeg, Song, and Scaramuzza]{heeg2025icra}
J.~Heeg, Y.~Song, and D.~Scaramuzza.
\newblock Learning quadrotor control from visual features using differentiable
  simulation, 2025.
\newblock URL \url{https://arxiv.org/abs/2410.15979}.

\bibitem[Belkhale et~al.(2021)Belkhale, Li, Kahn, McAllister, Calandra, and
  Levine]{belkhale2021ral}
S.~Belkhale, R.~Li, G.~Kahn, R.~McAllister, R.~Calandra, and S.~Levine.
\newblock Model-based meta-reinforcement learning for flight with suspended
  payloads.
\newblock \emph{IEEE Robotics and Automation Letters}, 6:\penalty0 1471--1478,
  02 2021.
\newblock \doi{10.1109/LRA.2021.3057046}.

\bibitem[Lee et~al.(2026)Lee, Rathod, Goel, Stecklein, and Tabib]{lee2025icra}
J.~Lee, A.~Rathod, K.~Goel, J.~Stecklein, and W.~Tabib.
\newblock Quadrotor navigation using reinforcement learning with privileged
  information.
\newblock \emph{2026 IEEE International Conference on Robotics and Automation
  (ICRA)}, May 2026.
\newblock \doi{10.48550/arXiv.2509.08177}.
\newblock URL \url{https://arxiv.org/pdf/2509.08177}.

\bibitem[Li et~al.(2017)Li, Liu, Zhang, Yeung, and Shen]{li2017arxiv}
S.~Li, T.~Liu, C.~Zhang, D.-Y. Yeung, and S.~Shen.
\newblock Learning unmanned aerial vehicle control for autonomous target
  following, 2017.
\newblock URL \url{https://arxiv.org/abs/1709.08233}.

\bibitem[Kaufmann et~al.(2022)Kaufmann, Bauersfeld, and
  Scaramuzza]{kaufmann2022icra}
E.~Kaufmann, L.~Bauersfeld, and D.~Scaramuzza.
\newblock A benchmark comparison of learned control policies for agile
  quadrotor flight.
\newblock In \emph{2022 International Conference on Robotics and Automation
  (ICRA)}, pages 10504--10510, 2022.
\newblock \doi{10.1109/ICRA46639.2022.9811564}.

\bibitem[Ji et~al.(2025)Ji, Wang, and He]{ji2025bicopter}
S.~Ji, Y.~Wang, and X.~He.
\newblock Flipping an upright-inverted bimodal bicopter uav: Attitude control
  and optimization.
\newblock \emph{IEEE/ASME Transactions on Mechatronics}, 30\penalty0
  (6):\penalty0 5063--5073, 2025.
\newblock \doi{10.1109/TMECH.2025.3549030}.

\bibitem[Mahony et~al.(2012)Mahony, Kumar, and Corke]{mahony2021multirotor}
R.~Mahony, V.~Kumar, and P.~Corke.
\newblock Multirotor aerial vehicles: Modeling, estimation, and control of
  quadrotor.
\newblock \emph{IEEE Robotics \& Automation Magazine}, 19\penalty0
  (3):\penalty0 20--32, 2012.
\newblock \doi{10.1109/MRA.2012.2206474}.

\bibitem[Huber(1964)]{huber1992loss}
P.~J. Huber.
\newblock {Robust Estimation of a Location Parameter}.
\newblock \emph{The Annals of Mathematical Statistics}, 35\penalty0
  (1):\penalty0 73 -- 101, 1964.
\newblock \doi{10.1214/aoms/1177703732}.
\newblock URL \url{https://doi.org/10.1214/aoms/1177703732}.

\bibitem[Schulman et~al.(2017)Schulman, Wolski, Dhariwal, Radford, and
  Klimov]{schulman2017ppo}
J.~Schulman, F.~Wolski, P.~Dhariwal, A.~Radford, and O.~Klimov.
\newblock Proximal policy optimization algorithms.
\newblock \emph{CoRR}, abs/1707.06347, 2017.
\newblock URL \url{http://arxiv.org/abs/1707.06347}.

\bibitem[Watterson and Kumar(2020)]{watterson2020rr}
M.~Watterson and V.~Kumar.
\newblock Control of quadrotors using the hopf fibration on so(3).
\newblock In N.~M. Amato, G.~Hager, S.~Thomas, and M.~Torres-Torriti, editors,
  \emph{Robotics Research}, pages 199--215, Cham, 2020. Springer International
  Publishing.
\newblock ISBN 978-3-030-28619-4.

\bibitem[Spitzer and Michael(2020)]{spitzer2020corr}
A.~Spitzer and N.~Michael.
\newblock Rotational error metrics for quadrotor control.
\newblock \emph{CoRR}, abs/2011.11909, 2020.
\newblock URL \url{https://arxiv.org/abs/2011.11909}.

\bibitem[Nurlanov(2024)]{nurlanov2024so3}
Z.~Nurlanov.
\newblock $so(3)$ transformations and jacobian of extended log map.
\newblock Technical report, Legged Robotics (Kindr), 2024.
\newblock URL \url{https://github.com/nurlanov-zh/so3_log_map}.

\bibitem[Bertsekas(2016)]{bertsekas2016nonlinear}
D.~Bertsekas.
\newblock \emph{Nonlinear Programming}.
\newblock Athena scientific optimization and computation series. Athena
  Scientific, 2016.
\newblock ISBN 9781886529052.
\newblock URL \url{https://books.google.com/books?id=rC1EEAAAQBAJ}.

\bibitem[Bradbury et~al.(2018)Bradbury, Frostig, Hawkins, Johnson, Katariya,
  Leary, Maclaurin, Necula, Paszke, Vander{P}las, Wanderman-{M}ilne, and
  Zhang]{jax2018github}
J.~Bradbury, R.~Frostig, P.~Hawkins, M.~J. Johnson, Y.~Katariya, C.~Leary,
  D.~Maclaurin, G.~Necula, A.~Paszke, J.~Vander{P}las, S.~Wanderman-{M}ilne,
  and Q.~Zhang.
\newblock {JAX}: composable transformations of {P}ython+{N}um{P}y programs,
  2018.
\newblock URL \url{http://github.com/jax-ml/jax}.

\bibitem[Cano~Lopes et~al.(2018)Cano~Lopes, Ferreira, da~Silva~Simões, and
  Luna~Colombini]{lopes2018ppo}
G.~Cano~Lopes, M.~Ferreira, A.~da~Silva~Simões, and E.~Luna~Colombini.
\newblock Intelligent control of a quadrotor with proximal policy optimization
  reinforcement learning.
\newblock In \emph{2018 Latin American Robotic Symposium, 2018 Brazilian
  Symposium on Robotics (SBR) and 2018 Workshop on Robotics in Education
  (WRE)}, pages 503--508, 2018.
\newblock \doi{10.1109/LARS/SBR/WRE.2018.00094}.

\bibitem[Milnor(1978)]{milnor1978hbt}
J.~Milnor.
\newblock Analytic proofs of the "hairy ball theorem" and the brouwer fixed
  point theorem.
\newblock \emph{The American Mathematical Monthly}, 85\penalty0 (7):\penalty0
  521--524, 1978.
\newblock ISSN 00029890, 19300972.
\newblock URL \url{http://www.jstor.org/stable/2320860}.

\bibitem[Lyons(2022)]{lyons2022hopf}
D.~W. Lyons.
\newblock An elementary introduction to the hopf fibration, 2022.
\newblock URL \url{https://arxiv.org/abs/2212.01642}.

\bibitem[Ganesh et~al.(2025)Ganesh, Khidkikar, Velarde, Cox, Barngrover, and
  Michael]{edgeos}
V.~Ganesh, S.~Khidkikar, D.~Velarde, J.~Cox, C.~Barngrover, and N.~Michael.
\newblock Edge{OS}: A high-performance middleware framework for autonomous
  robotics.
\newblock
  \url{https://shield.ai/hivemind-edgeos-a-game-changer-for-autonomous-robotics/},
  2025.

\bibitem[developers(2021)]{onnxruntime}
O.~R. developers.
\newblock Onnx runtime.
\newblock \url{https://onnxruntime.ai/}, 2021.
\newblock Version: 1.20.1.

\bibitem[Mellinger(2012)]{mellinger2012thesis}
D.~Mellinger.
\newblock \emph{Trajectory generation and control for quadrotors}.
\newblock PhD thesis, 2012.
\newblock URL
  \url{https://www.proquest.com/dissertations-theses/trajectory-generation-control-quadrotors/docview/1018692309/se-2}.
\newblock Copyright - Database copyright ProQuest LLC; ProQuest does not claim
  copyright in the individual underlying works; Last updated - 2023-03-03.

\end{thebibliography}
\ifthenelse{\equal{\arxivmode}{true}}
{\clearpage}
{}
\appendix

\section{Topological Foundations of the Hopf Fibration-Based Control Algorithm}

\begin{wrapfigure}{r}{0.5\textwidth}
    \ifthenelse{\equal{\arxivmode}{true}}
    {\vspace{-5pt}}
    {\vspace{-10pt}}
    \centering
    \includegraphics[width=0.49\textwidth]{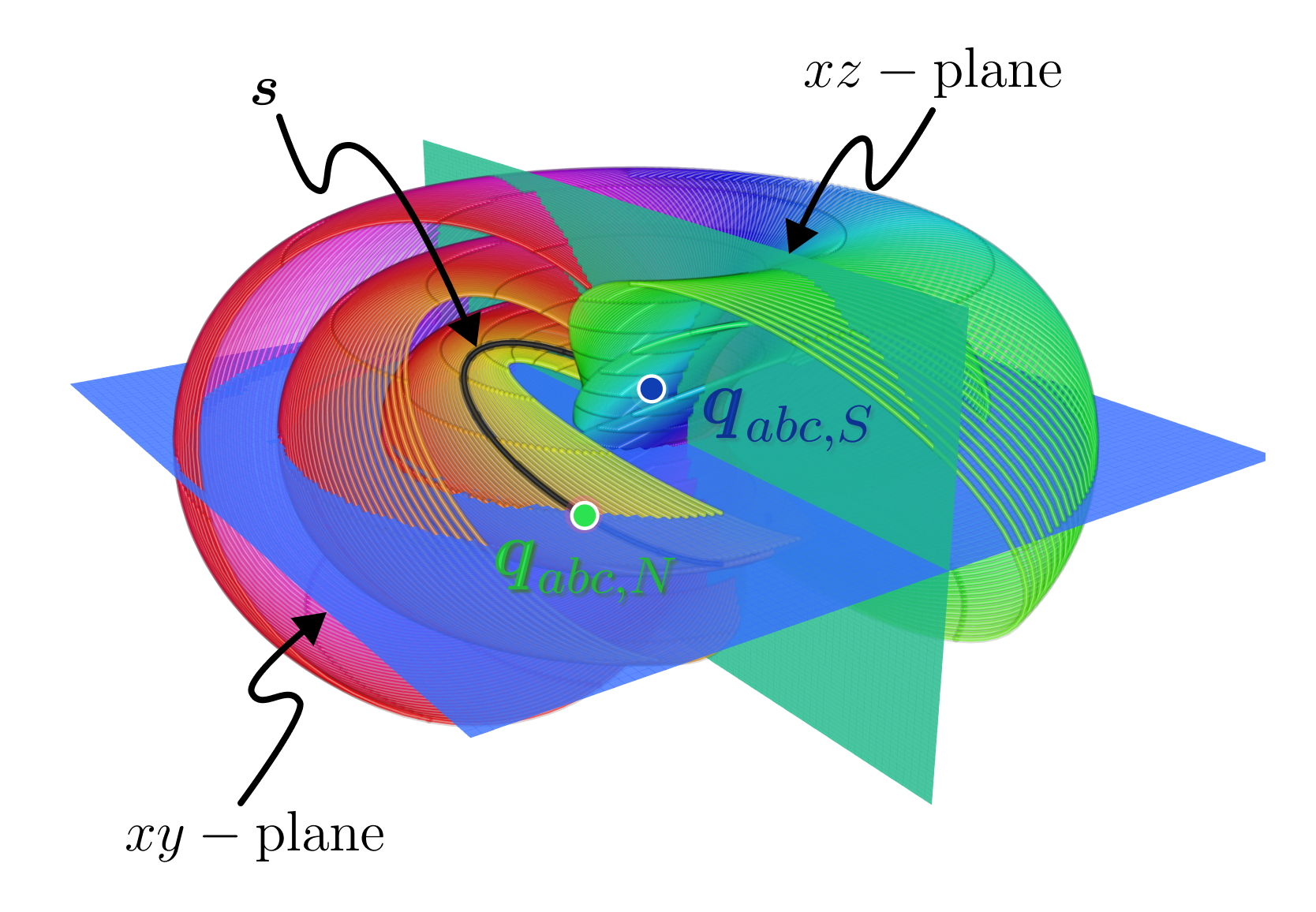}
    \caption{Visualization of the HFCA coordinate charts through stereographic
    projection.}
    \label{fig:hopf_intuition}
    \ifthenelse{\equal{\arxivmode}{true}}
    {\vspace{-5pt}}
    {\vspace{-15pt}}
\end{wrapfigure}

Modeling a quadrotor as a differentially flat hybrid system with
flat variables $\eta \in \{\pm1\}, (\bm{r}, \psi)$, is possible because thrust is
aligned with $\eta\hat{\mathbf{b}}_3$, the product of thrust posture and the
body $z$-axis.
This alignment allows the attitude to be decomposed into a thrust direction
$\bm{s} = \eta \hat{\mathbf{b}}_3 \in \Sp^2$ and a residual
rotation about this axis, $\psi \in \Sp^1$, which represents yaw.
However, this decoupling attempts to assign a globally consistent
yaw direction in $\Sp^1$ to every thrust direction in $\Sp^2$.
Since $\Sp^2$ cannot be covered by a single smooth coordinate chart,
requiring at least two charts with non-affine coordinate transformations,
this global assignment breaks down, yielding a singularity where
chart definitions conflict (cf. hairy ball theorem~\citep{milnor1978hbt}).

This topology is described by the Hopf fibration $\Sp^1 \hookrightarrow \Sp^3
\xrightarrow{\bm{p}} \Sp^2,$ where $\Sp^3$, viewed as the space of unit
quaternions representing attitude, is projected to thrust directions by the
Hopf map~\citep{lyons2022hopf}
\begin{equation}
    \bm{p}:\Sp^3 \rightarrow \Sp^2, \qquad
    \bm{q} \mapsto \text{Im}(\bm{q} \otimes \mathbf{k} \otimes\overline{\bm{q}}).
\end{equation}
For a given thrust direction $\bm{s} \in \Sp^2$, the fiber $\bm{p}^{-1}(\bm{s})$
represents all attitudes that have body $z$-axis equal to the thrust direction
$\bm{s}$. The fiber bundle
structure informs us that this fiber is circular.
Specifically, we can observe the fiber is invariant to yaw rotation about the body
$z$-axis, since for all $\bm{q} \in \bm{p}^{-1}(\bm{s}), \psi \in \Sp^1$ we have
\begin{equation}
    \bm{p}(\bm{q} \otimes e^{\psi \mathbf{k}}) =
    \text{Im}(\bm{q} \otimes e^{\psi\mathbf{k}} \otimes \mathbf{k}
        \otimes e^{-\psi\mathbf{k}} \otimes \overline{\bm{q}}) =
    \text{Im}(\bm{q} \otimes \mathbf{k} \otimes \overline{\bm{q}}) =
    \bm{p}(\bm{q}),
\end{equation}
hence right multiplication by $e^{\psi\mathbf{k}}$ parameterizes the fiber,
thus $\bm{q} \otimes e^{\psi\mathbf{k}} \in \bm{p}^{-1}(\bm{s})$.

In this context the Hopf Fibration-Based Control Algorithm (HFCA), first
presented by \citet{watterson2020rr}, expanded to the bidirectional thrust case by
\citet{watterson2020iser}, and re-introduced with modifications for better
inversion performance in~\cref{sec:HFCA}, maps a desired yaw $\psi \in \Sp^1$,
and thrust direction $\bm{s} \in \Sp^2$, to desired attitude in two steps.
First, a point along the fiber $\bm{p}^{-1}(\bm{s})$ is selected to map
$\mathbf{k}$ to $\eta\bm{s}$. Second, the yaw angle is incorporated as a
rotation about the $z$-axis, represented as $e^{\frac{\psi}{2}\mathbf{k}}$.

As noted, two charts are necessary to define this selection globally, defined in
\cref{eq:hopf_maps}. The primary chart $\bm{q}_{abc, N}$ is drawn from the
intersection of the $xy$-plane with the fiber, hence the rotation is minimal and
a tilt without twist. The secondary chart $\bm{q}_{abc, S}$ is constructed from
the intersection of the $xz$-plane with the fiber. Since the rotation is
generally not minimal, additional twist in yaw can occur. Such is minimized by
shifting the yaw reference to smoothen the transition from the primary to
secondary chart. The desired attitude is then the composition of the chart
derived quaternion with the yaw rotation.

The charts $\bm{q}_{abc, N}$ and $\bm{q}_{abc, S}$ are for $\Sp^2$. Thus, which
chart is selected only depends on the desired body $z$-axis,
$\hat{\mathbf{b}}_{3,d} = [a,b,c] \in \Sp^2$. Because the North chart
$\bm{q}_{abc, N}$ and South chart $\bm{q}_{abc, S}$ are
defined on $\Sp^2 \setminus \{(0,0,-1)\}$ and $\Sp^2 \setminus \{(0,0,1)\}$
respectively, we avoid their singularities by choosing to switch at the equator
($c = 0$) with a hysteresis to ensure the active chart is defined in the
currently occupied hemisphere, as in \citep{watterson2020iser}.

As a measure against the twist introduced by the secondary chart $\bm{q}_{abc, S}$
we consider
\begin{equation}
    \bm{q}_{abc, N}\otimes\bm{q}_\text{yaw}(\psi_N) = \bm{q}_{abc, S} \otimes \bm{q}_\text{yaw}(\psi_S)
\end{equation}
Left-multiplying by $\bm{q}_{abc, S}$ gives
\begin{equation}
    \overline{\bm{q}}_{abc, S}\otimes\bm{q}_{abc, N}=\bm{q}_\text{yaw}(\psi_S - \psi_N),
\end{equation}
substituting the chart definitions in~\cref{eq:hopf_maps} yields
\begin{equation}
    \bm{q}_\text{yaw}(\psi_S - \psi_N) = \frac{1}{\sqrt{1-c^2}}[-b, 0, 0, -a]^\top.
\end{equation}
Comparing with yaw quaternion definition in~\cref{sec:HFCA}, $\psi_S - \psi_N =
2\operatorname{atan2}(a,b)$. Hence, continuity between charts is maintained by
setting $\psi_S$ equal to the nominal yaw reference $\psi_N$ plus a saved offset
$2\operatorname{atan2}(a,b)$, provided that $\operatorname{atan2}(a,b)$ remains
unchanged in the South chart. If $\operatorname{atan2}(a,b)$ does change, then
$\psi_S$ may gradually diverge from $\psi_N$. Updating
$\operatorname{atan2}(a,b)$ while in the South chart would improve $\psi_S$
tracking of the desired $\psi_N$, but would become singular at inverted hover
$[a,b,c]=[0,0,-1]$, proving impractical and reflecting the absence of a globally
nonsingular coordinate chart on $\Sp^2$.

\section{Hardware Deployment}

\subsection{Experimental Platform}
\begin{figure}
  \begin{center}
    \subfloat[Top-down view of the aerial platform]{\label{sfig:top-down-drone}\includegraphics[height=5cm]{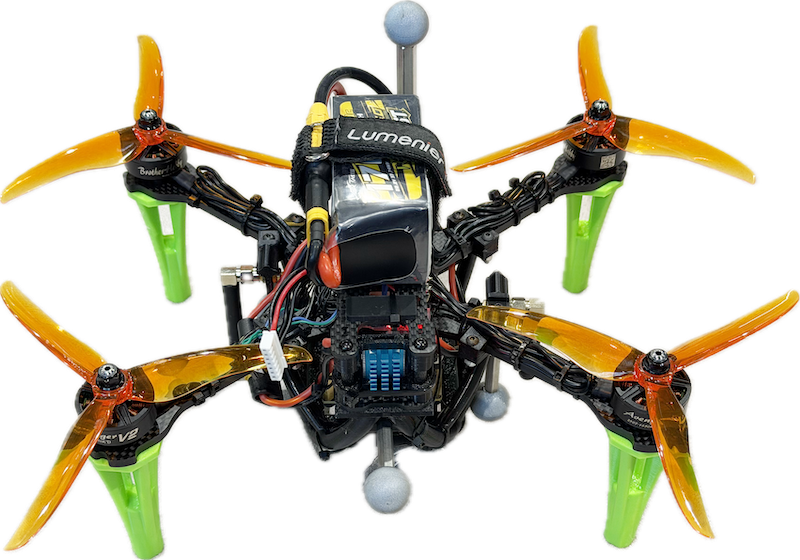}}\hfill %
    \subfloat[Upside-down view of the aerial platform]{\label{sfig:upside-down-drone}\includegraphics[height=5cm]{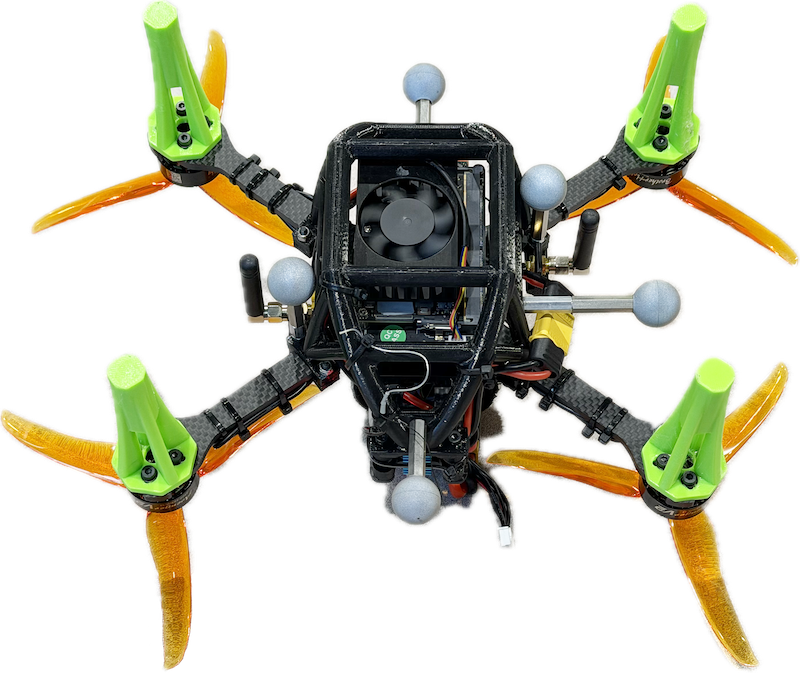}}\\
    \subfloat[Nominal circle trajectory]{\label{sfig:nominal-circle}\includegraphics[height=4.4cm]{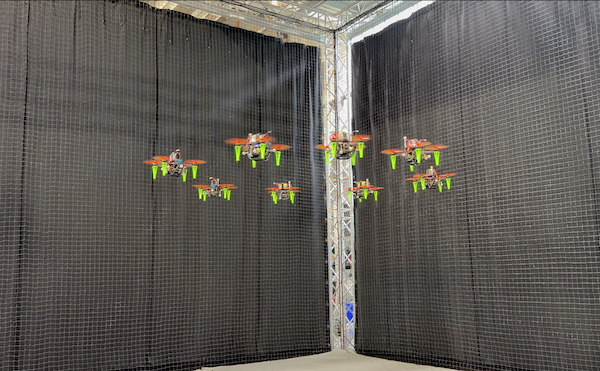}}\hfill %
    \subfloat[Inverted circle trajectory]{\label{sfig:inverted-circle}\includegraphics[height=4.4cm]{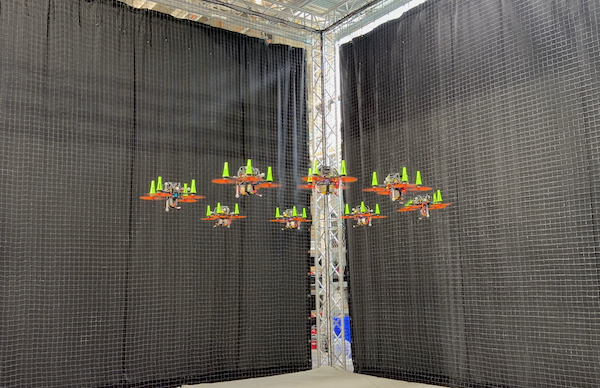}}\\
  \end{center}
  \caption{\label{fig:hardware} Custom quadrotor platform
    developed for the hardware experiments.~\protect\subref{sfig:top-down-drone}
    illustrates the top view of the drone with 3D printed landing
    legs in neon green and~\protect\subref{sfig:upside-down-drone} illustrates
    the underside of the drone with the black 3D printed cage to protect the
    NVIDIA Orin NX onboard computer and carrier board.~\protect\subref{sfig:nominal-circle}
    illustrates one hardware experiment where the platform
    follows a circle trajectory and~\protect\subref{sfig:inverted-circle}
    illustrates the same hardware experiment after the platform has inverted
    and executes another circle trajectory.}
\end{figure}
The quadrotor is a custom-built 6'' platform (shown
in~\cref{fig:hardware}) created with the Lumenier QAV-R 2 Freestyle
Quadcopter frame, TBS Lucid H7 flight controller, TBS Lucid 60A AM32
4-in-1 ESC, and an NVIDIA Orin NX (16~GB) with Seeedstudio A603 carrier
board. The motors are Brotherhobby
Avenger V2 2507-1850KV with T-Motor T6143 propellers powered using a
6S Tattu 1700~mAh battery. The system ran Ubuntu 22.04 using Jetpack
6.2 with communications between the flight controller and carrier
board running over UART. A cage was 3D printed to protect
the Orin NX and carrier board using black TPU for AMS filament. The
cage was printed with 4 wall loops and 50\% infill. Neon green legs
were also 3D printed from the same material (100\% infill).  The TPU
for AMS was chosen for its ability to absorb an impact and the
strength to withstand breaking. Six reflective markers
(\SI{19}{\milli\meter}) were mounted to the frame so that a minimum of
four are viewable by the motion tracking system for any given
orientation of the platform.

\subsection{Implementation Details}
The system ran Betaflight firmware. The middleware used was ROS2
Humble and Shield AI's EdgeOS~\cite{edgeos}.  All control
algorithms ran onboard the Orin NX and DShot commands were sent over
UART to the flight controller at a rate of \SI{1}{\kilo\hertz}. Policy
inference operated at a rate of \SI{50}{\hertz} using ONNX Runtime's TensorRT
Execution Provider~\citep{onnxruntime}, position control at
\SI{100}{\hertz}, and orientation control at \SI{1}{\kilo\hertz}.

\subsection{System Identification\label{ssec:sysid}}

\begin{figure}
  \begin{center}
    \subfloat[Steady-state thrust model]{\label{sfig:ss_thrust_model}\includegraphics[height=4.5cm]{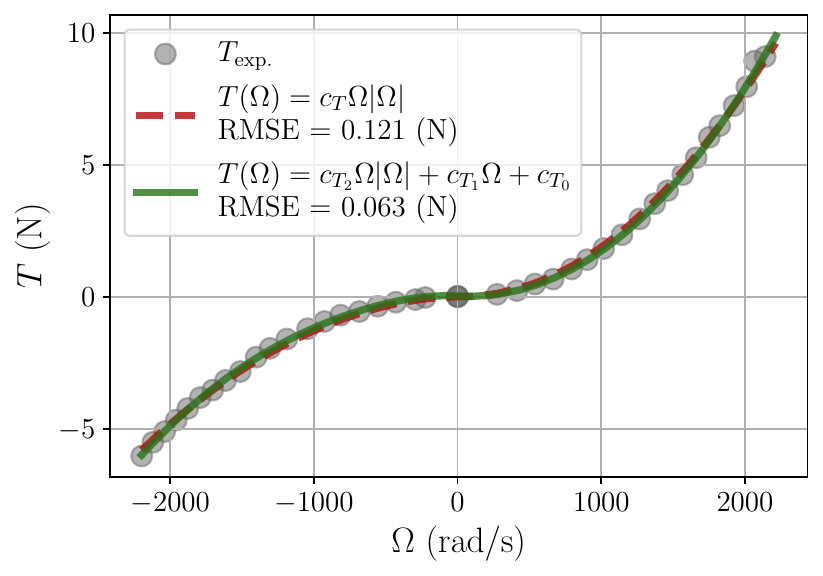}}\hfill %
    \subfloat[Steady-state torque model]{\label{sfig:ss_torque_model}\includegraphics[height=4.5cm]{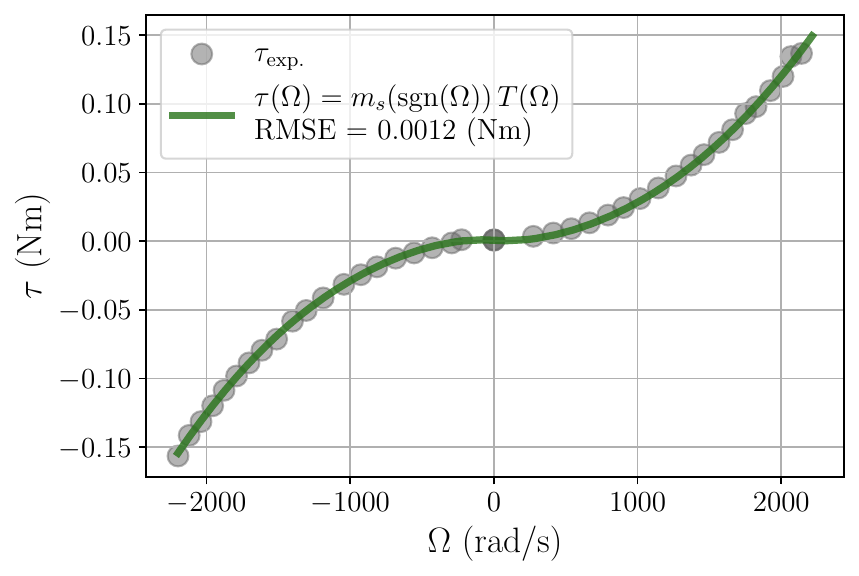}}\\
    \subfloat[$(+)$ transient model]{\label{sfig:t_pos_model}\includegraphics[height=4.5cm]{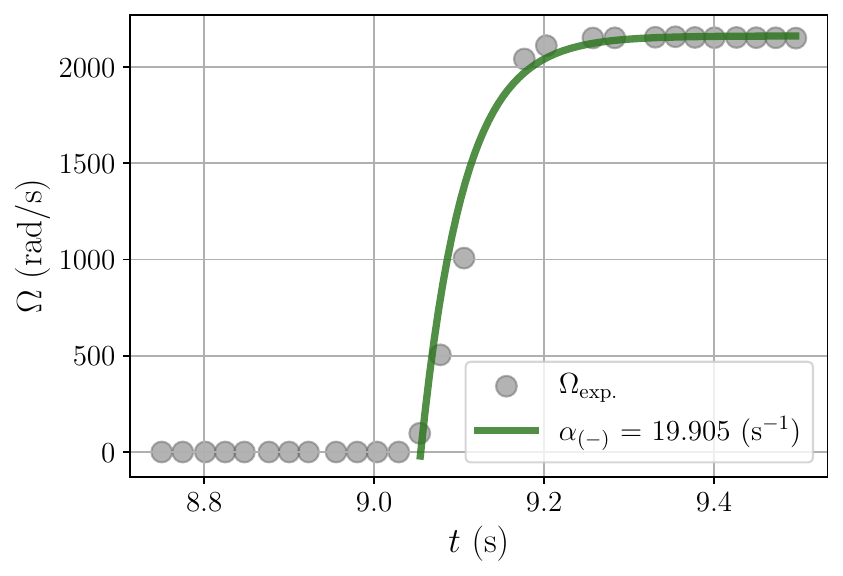}}\hfill %
    \subfloat[$(-)$ transient model]{\label{sfig:t_neg_model}\includegraphics[height=4.5cm]{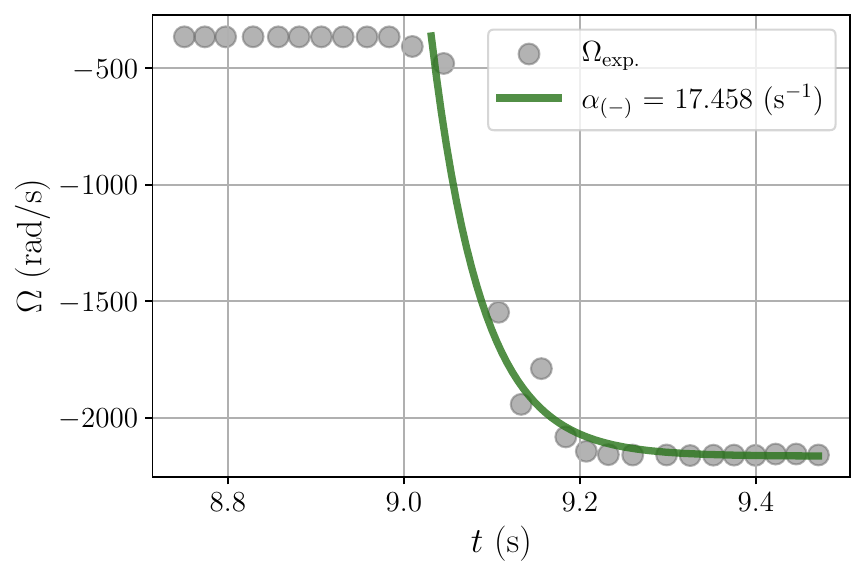}}\\
    \subfloat[$(+){\rightarrow}(-)$ transient model]{\label{sfig:t_pos_neg_model}\includegraphics[height=4.5cm]{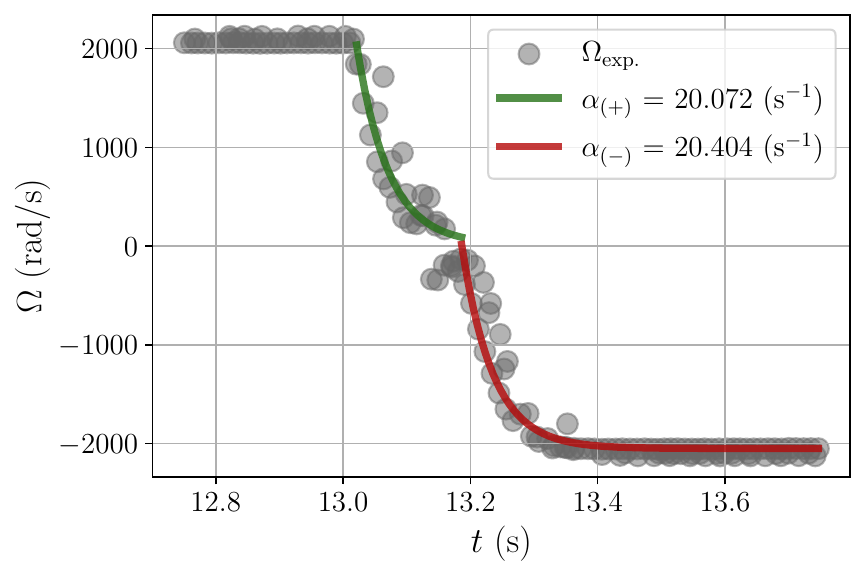}}\hfill %
    \subfloat[$(-){\rightarrow}(+)$ transient model]{\label{sfig:t_neg_pos_model}\includegraphics[height=4.5cm]{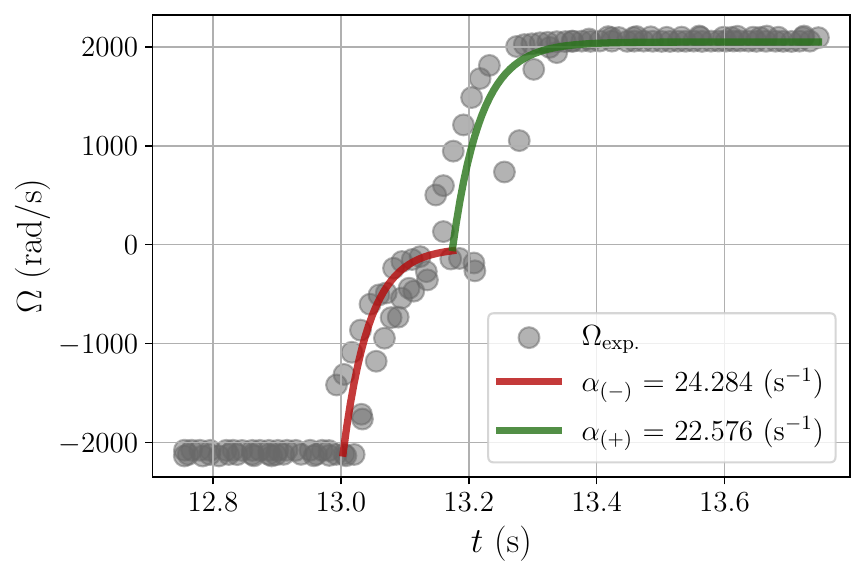}}\\
  \end{center}
  \caption{\label{fig:thrust_models} Thrust and torque model fit to experimental
    data collected using the Tyto Robotics Series 1585 drone thrust stand and RC
    control board.
    ~\protect\subref{sfig:ss_thrust_model} illustrates how
    asymmetric propellers generate significantly more thrust in the $(+)$
    operating regime (about 2$\times$) compared to the $(-)$ operating
    regime.
    ~\protect\subref{sfig:ss_torque_model} illustrates the torque 
    varying across operating regimes due to propeller efficiency
    asymmetry (i.e., the moment scale is larger in the (-) regime
      because it is less efficient).
    ~\protect\subref{sfig:t_pos_model} illustrates the first-order
    model fit to the $(+)$ operating regime of the propeller.
    ~\protect\subref{sfig:t_neg_model} illustrates the first-order
    model fit to the $(-)$ operating regime of the propeller.
    ~\protect\subref{sfig:t_pos_neg_model} illustrates the piecewise first-order
    model fit to the $(+){\rightarrow}(-)$ operating regime transition, and the
    apparent stochasticity and asymmetry across regimes.
    ~\protect\subref{sfig:t_neg_pos_model} illustrates the piecewise first-order
    model fit to the $(-){\rightarrow}(+)$ operating regime transition, and the
    apparent stochasticity and asymmetry across regimes.}
\end{figure}

The steady-state thrust and moment scale models were fit via least squares from
data collected using the Tyto Robotics Series 1585 Drone Thrust Stand and RC
control board (to send DShot commands and measure the response). Results for
steady-state model fitting are shown in~\cref{sfig:ss_thrust_model,sfig:ss_torque_model}.
The transient model parameters were identified from
step-response experiments: rise times were extracted within each operating
regime, and during transitions between them. The $(+){\rightarrow}(-)$ and
$(-){\rightarrow}(+)$ transition points were identified manually from the motor
rate response, with results shown in~\cref{sfig:t_pos_model,sfig:t_neg_model,sfig:t_pos_neg_model,sfig:t_neg_pos_model} with a corresponding simulation of
our transient thrust model (\cref{sec:model}) overlaying the experimental data
for the same desired motor rates.

The mass of the quadrotor and its motors was measured directly on the hardware,
while the location of the center of mass ($CM$) relative to the geometric
center $(GC)$, and the inertia matrix were estimated in a computer-aided design
(CAD) program. Estimating the center of mass offset was crucial for hardware
deployment. Following the approach of \citet{mellinger2012thesis}, the center of
mass in $\mathcal{W}$ can be defined relative to the geometric center by the
relation $\bm{r}_{CM} = \bm{r}_{GC} + \bm{R}\bm{r}_{\text{off}}$, where
$\bm{r}_\text{off}$ is a vector in $\mathcal{B}$ that defines the offset from
the geometric center to the center of mass and $\bm{R}$ defines the rotation
matrix from $\mathcal{B}$ to $\mathcal{W}$. Using this, the equations of motion
are modified to~\citep{mellinger2012thesis}:
\begin{align}
    m \left( \ddot{\bm{r}}_\text{GC}
    + [\bm{\dot{\omega}}]_\times \bm{r}_{\text{off}}
    + [\bm{\omega}]_\times^2 \bm{r}_{\text{off}} \right)
    &= -mg\hat{\mathbf{e}}_3 + f_c\hat{\mathbf{b}}_3, \\
    \bm{\mathcal{I}}_{CM} \dot{\bm{\omega}} &=
    -[\bm{\omega}]_\times \bm{\mathcal{I}}_{CM} \bm{\omega}
    + \bm{\tau}
    - \bm{r}_\text{off} \times \left[0, 0, f_c \right]^\top,
\end{align}
where we use $[\mathbf{\cdot}]_\times$ to denote skew-symmetric matrices.
Although the simulator does not account for $CM$ offsets, the controller onboard
the robot must reject its effect; thus, we modify the attitude control law to:
\begin{equation}
    \bm{\tau} = - \bm{J}_R^{-\top}(\bm{e}_R) \bm{K_R} \bm{e_R} - \bm{K_\omega} \bm{e_\omega} + \bm{r}_{\text{off}} \times \left[0, 0, f_c \right]^\top.
\end{equation}
Lastly, the policy inference and communication delay was characterized using
a throttled ROS2 logger to print system time differences between observation
concatenation and application to HFCA implementation.

\section{Training \& Simulation}

We train all policies in our JAX-based simulator, {\small\texttt{acrorl}},
using an RTX 4090. For evaluation and deployment, we use the mean of the learned
stochastic policy, and introduce an inversion flag $\mu$ to switch from
trajectory tracking geometric control to inversion on the hardware platform.
During training, environments are initialized from randomized orientations,
angular rates, positions, and velocities to improve robustness.

The measured inference and communication delay from the hardware deployment
framework is modeled in simulation to aid the sim2real transfer. Training is
completed within 9 minutes for both policies with 2048 parallel environments and
750 training epochs.~\cref{tab:ppo_params} reports the model architectures,
hyperparameters, and key environmental setup choices for both NtI and ItN training,
while~\cref{tab:reward_weights} reports the individual weights applied
to~\cref{eq:reward}. We also report the best performing minimum snap trajectory
parameters in~\cref{tab:min_snap}.

\begin{table}[htbp]
\centering
\footnotesize
\begin{tabular}{l l r}
\toprule
\textbf{Parameter} & \textbf{Description} & \textbf{Value} \\
\midrule
\multicolumn{3}{l}{\textit{Network Architecture}} \\
\midrule
Hidden layer dims      & MLP layer widths                        & $[512, 512]$ \\
Initial $\log \sigma$  & Initial log std of action distribution  & $\log(0.25)$ \\
Action history         & Number of past actions in observation   & $3$ \\
\midrule
\multicolumn{3}{l}{\textit{PPO Hyperparameters}} \\
\midrule
Learning rate          & Adam optimizer learning rate            & $3 \times 10^{-4}$ \\
$\gamma$               & Discount factor                         & $0.99$ \\
$\lambda$              & GAE decay parameter                     & $0.95$ \\
Clip $\epsilon$        & PPO clipping parameter                  & $0.2$ \\
Entropy coefficient    & Entropy regularization weight           & $0.01$ \\
Value function coef.   & Value loss weight                       & $0.5$ \\
Update epochs          & Gradient epochs per rollout             & $4$ \\
Minibatches            & Number of minibatches per update        & $20$ \\
\midrule
\multicolumn{3}{l}{\textit{Training Setup}} \\
\midrule
Num.\ environments     & Parallel rollout environments           & $2048$ \\
Num.\ epochs           & Total training epochs                   & $750$ \\
Episode duration       & Training episode length                 & \SI{3}{\second} \\
$dt_\text{env}$        & Simulation environment timestep         & \SI{0.02}{\second}\\
$dt_\text{dyn}$        & Simulation dynamics timestep            & \SI{0.001}{\second} \\
Delay                  & Simulated policy inference delay        & \SI{0.006}{\second} \\
\bottomrule
\end{tabular}
\vspace{0.4em}
\caption{PPO training hyperparameters and environment configuration.}
\label{tab:ppo_params}
\end{table}

\begin{table}[htbp]
\centering
\footnotesize
\begin{tabular}{lcc}
\toprule
\textbf{Reward term} & \textbf{ItN} & \textbf{NtI} \\
\midrule
Position, $w_{\bm{r}}$               & 5.000 & 5.000 \\
Orientation, $w_{\bm{g}_b}$             & 3.000 & 3.000 \\
Thrust posture, $w_{\tilde{\eta}}$         & 0.100 & 0.100 \\
Velocity, $w_{\bm{\dot{r}}}$               & 0.000 & 0.005 \\
Angular velocity, $w_{\bm{\omega}}$        & 0.750 & 0.200 \\
Action rate $w_{\bm{\dot{r}}_{\delta \kappa}}$  & 0.250 & 0.200 \\
\bottomrule
\end{tabular}
\vspace{0.4em}
\caption{Reward weights used for inverted-to-nominal (ItN) and nominal-to-inverted (NtI) training.}
\label{tab:reward_weights}
\end{table}

\begin{table}[htbp]
\centering
\footnotesize
\begin{tabular}{lc}
\toprule
\textbf{Parameter} & \textbf{Value}  \\
\midrule
Second waypoint $\delta z$      & \SI{0.45}{\meter}\\
$\eta$ schedule                 & (1.0, -1.0)\\
Trajectory duration             & \SI{2.0}{\second} \\
Segment durations               & (1.0, 1.0) \si{\second} \\
\bottomrule
\end{tabular}
\vspace{0.4em}
\caption{Tuned minimum snap trajectory parameters for best inversion performance.}
\label{tab:min_snap}
\end{table}


\end{document}